\documentclass[10pt,journal,twocolumn]{IEEEtran}
\usepackage{graphicx}
\usepackage{float}
\usepackage{amsmath}
\usepackage{amssymb}
\usepackage{subfigure}
\usepackage{amsfonts}
\usepackage{placeins}
\usepackage{algorithm}
\usepackage{algpseudocode}
\usepackage{epstopdf}
\usepackage{comment}
\usepackage{amssymb,amsmath,amsthm}
\usepackage{xcolor}
\usepackage[nocompress]{cite}
\usepackage{cite}
\usepackage{setspace}
\usepackage[compact]{titlesec}
\usepackage{tabularx, booktabs}
\usepackage[pagebackref=true,breaklinks=true,letterpaper=true,colorlinks,bookmarks=false]{hyperref}

\graphicspath{{figures/}}
\begin{document}

\title{Synthesis and Reconstruction of Fingerprints using Generative Adversarial Networks}
\author{Rafael Bouzaglo and Yosi~Keller
\IEEEcompsocitemizethanks{\IEEEcompsocthanksitem R. Bouzaglo \& Y. Keller are with the Faculty of Engineering, Bar-Ilan University,
E-mail:yosi.keller@gmail.com}\thanks{%
Manuscript received April 19, 2005; revised August 26, 2015.}}
\maketitle

\begin{abstract}
Deep learning-based models have been shown to improve the accuracy of
fingerprint recognition. While these algorithms show exceptional
performance, they require large-scale fingerprint datasets for training and
evaluation. In this work, we propose a novel fingerprint synthesis and
reconstruction framework based on the StyleGan2 architecture, to address the
privacy issues related to the acquisition of such large-scale datasets. We
also derive a computational approach to modify the attributes of the
generated fingerprint while preserving their identity. This allows for the
simultaneous synthesis of multiple different fingerprint images per finger.
In particular, we introduce the SynFing synthetic fingerprints dataset
consisting of 100K image pairs, each pair corresponding to the same
identity. The proposed framework was experimentally shown to outperform
contemporary state-of-the-art approaches for both fingerprint synthesis and
reconstruction. It significantly improved the realism of the generated
fingerprints, both visually and in terms of their ability to spoof
fingerprint-based verification systems. The code and fingerprints dataset
are publicly available: %
\url{https://github.com/rafaelbou/fingerprint-generator/}.
\end{abstract}


\IEEEdisplaynontitleabstractindextext
\IEEEpeerreviewmaketitle

\section{Introduction}

\label{sec:introduction}

Human fingerprints are one of the most common biometric attributes used for
authentication and identification, from border control identification to
payment authorization to the daily use of unlocking electronic devices
such as cellphones \cite{maltoni2009handbook, jain2010biometrics}. Current
state-of-the-art fingerprint recognition systems are mostly based on deep
learning models \cite{minaee2019biometrics, sundararajan2018deep}. While
these systems show exceptional performance, they require expensive
large-scale fingerprint datasets for training and evaluation. Furthermore,
collecting and sharing large-scale biometric datasets comes with inherent
risks and privacy concerns. For example, the National Institute of Standards
and Technology (NIST) recently discontinued several publicly available
datasets from its catalog due to privacy issues \cite{NIST:Catalog}.
\begin{figure}[th]
\subfigure[Original]{\includegraphics[width=0.25\columnwidth,height=0.25%
\columnwidth]{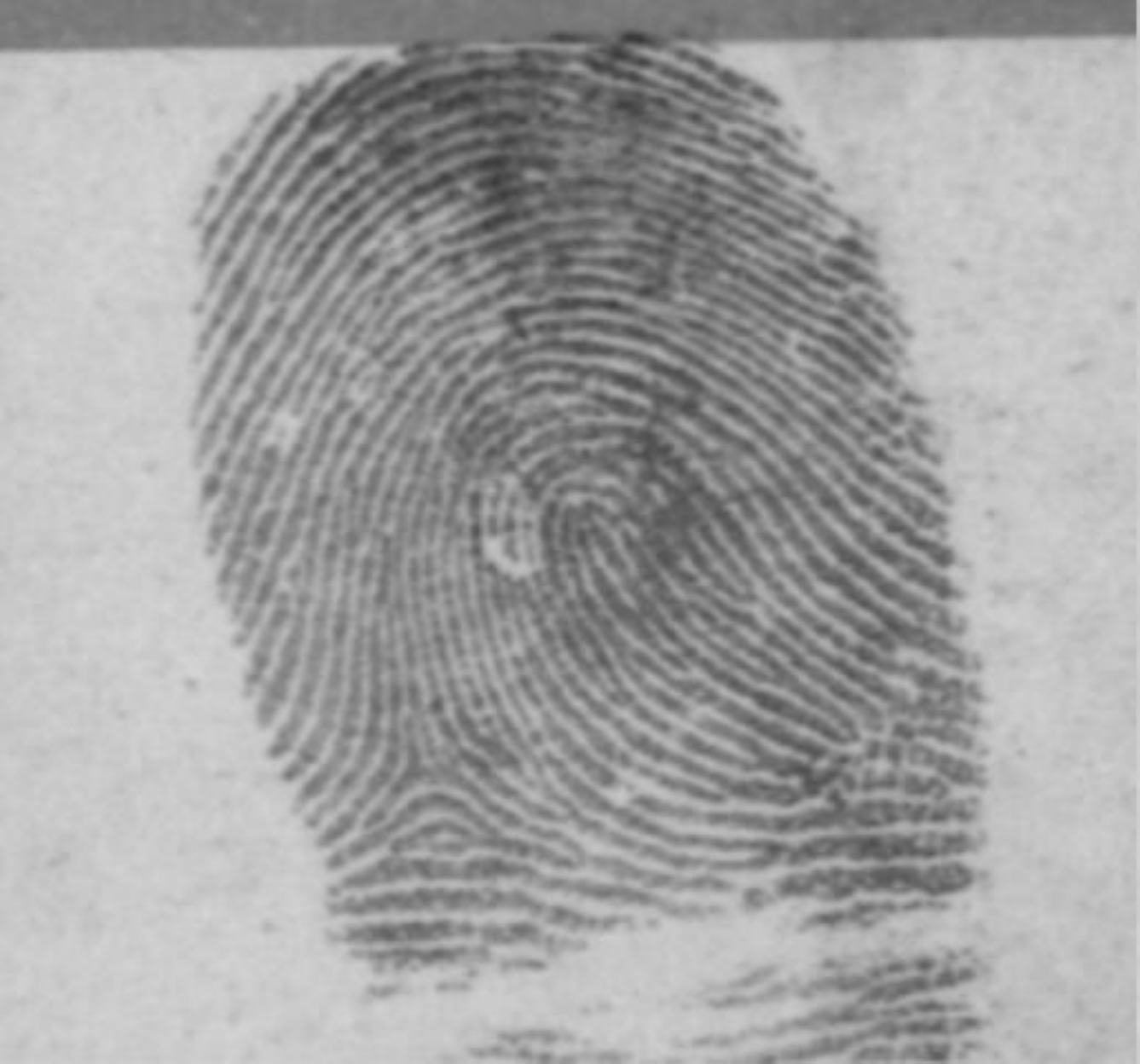}}\subfigure[Reconstructed]{%
\includegraphics[width=0.26\columnwidth,height=0.25%
\columnwidth]{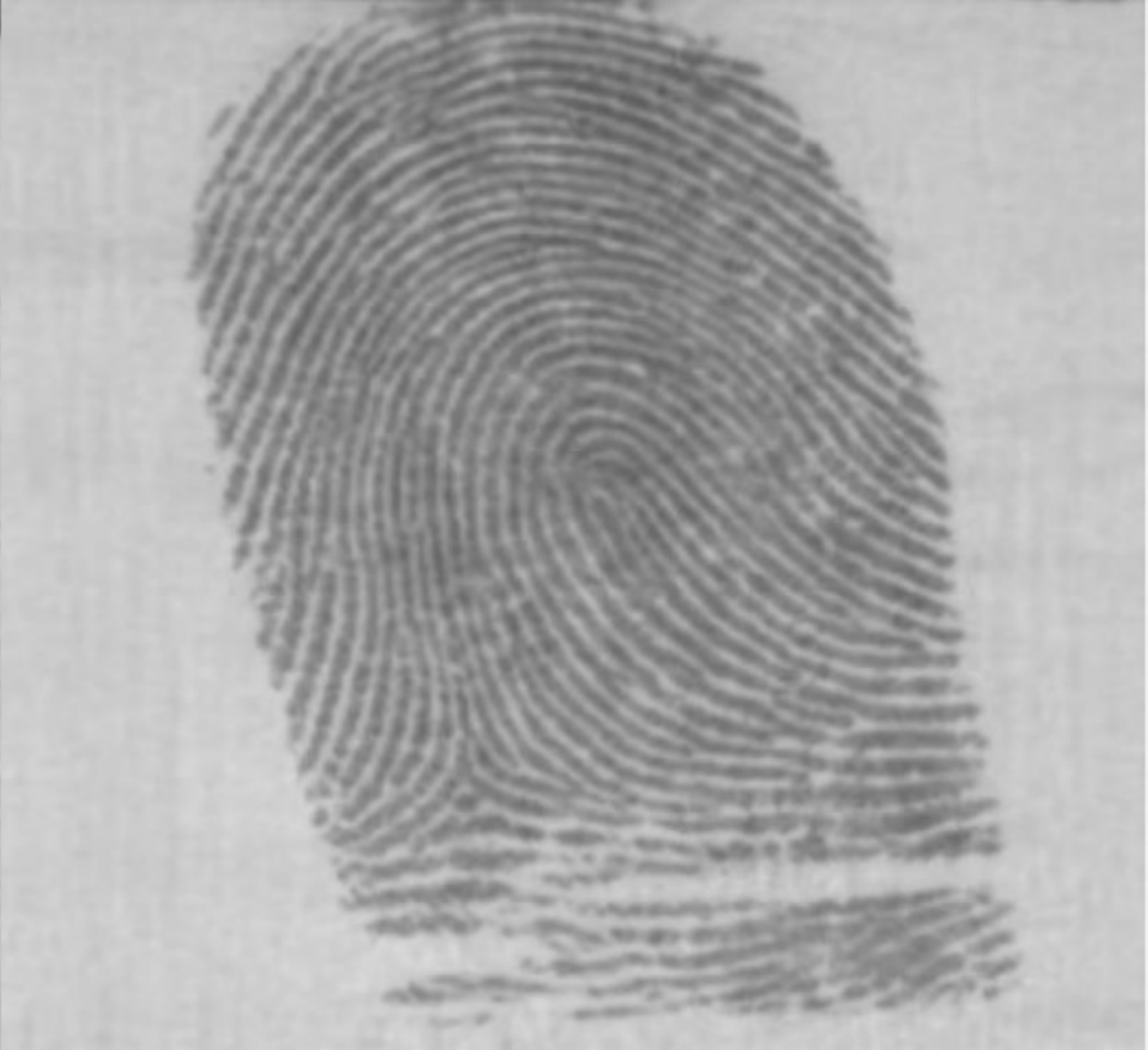}}\subfigure[Synthesized]{%
\includegraphics[width=0.25\columnwidth,height=0.25%
\columnwidth]{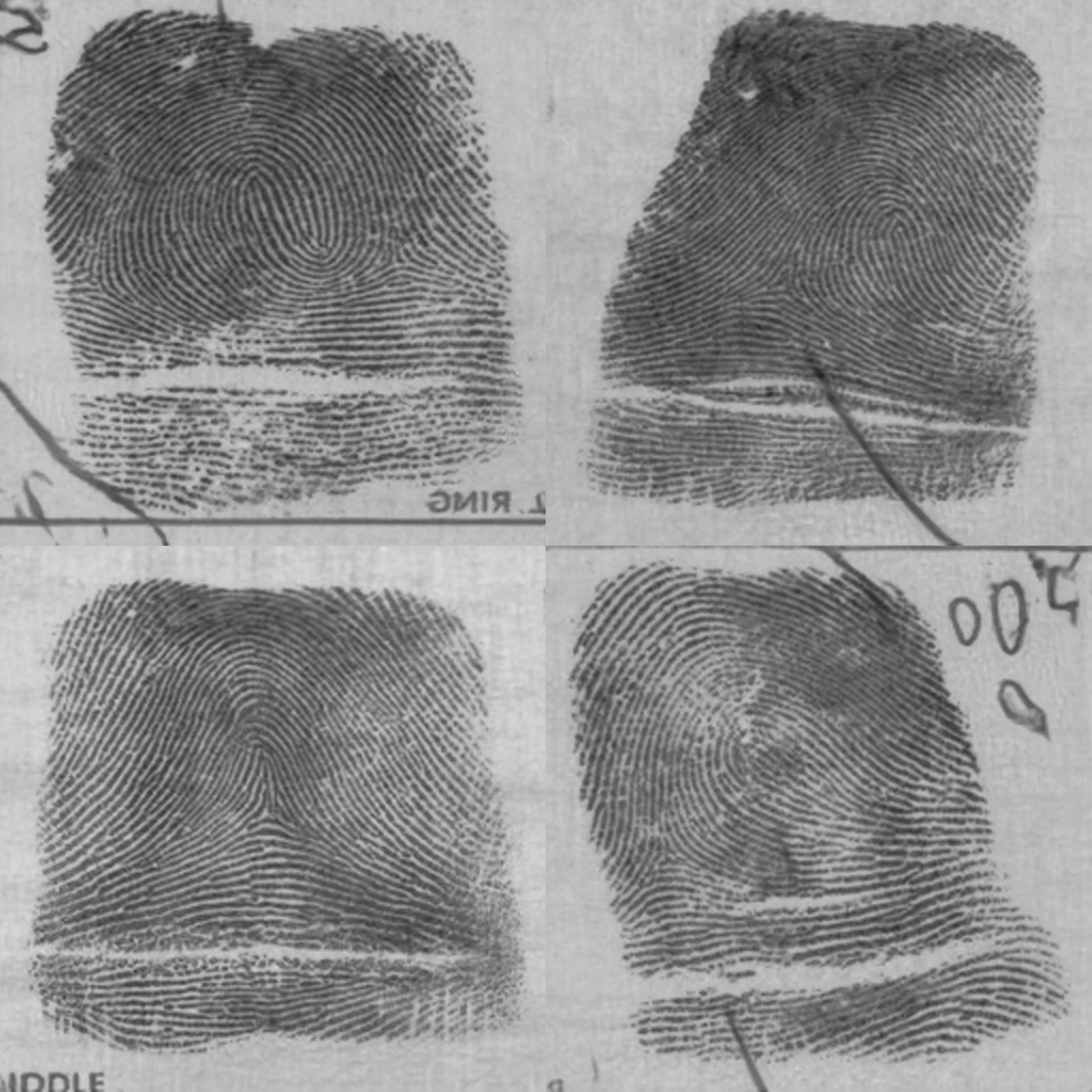}}\subfigure[Synthesized]{%
\includegraphics[width=0.25\columnwidth,height=0.25%
\columnwidth]{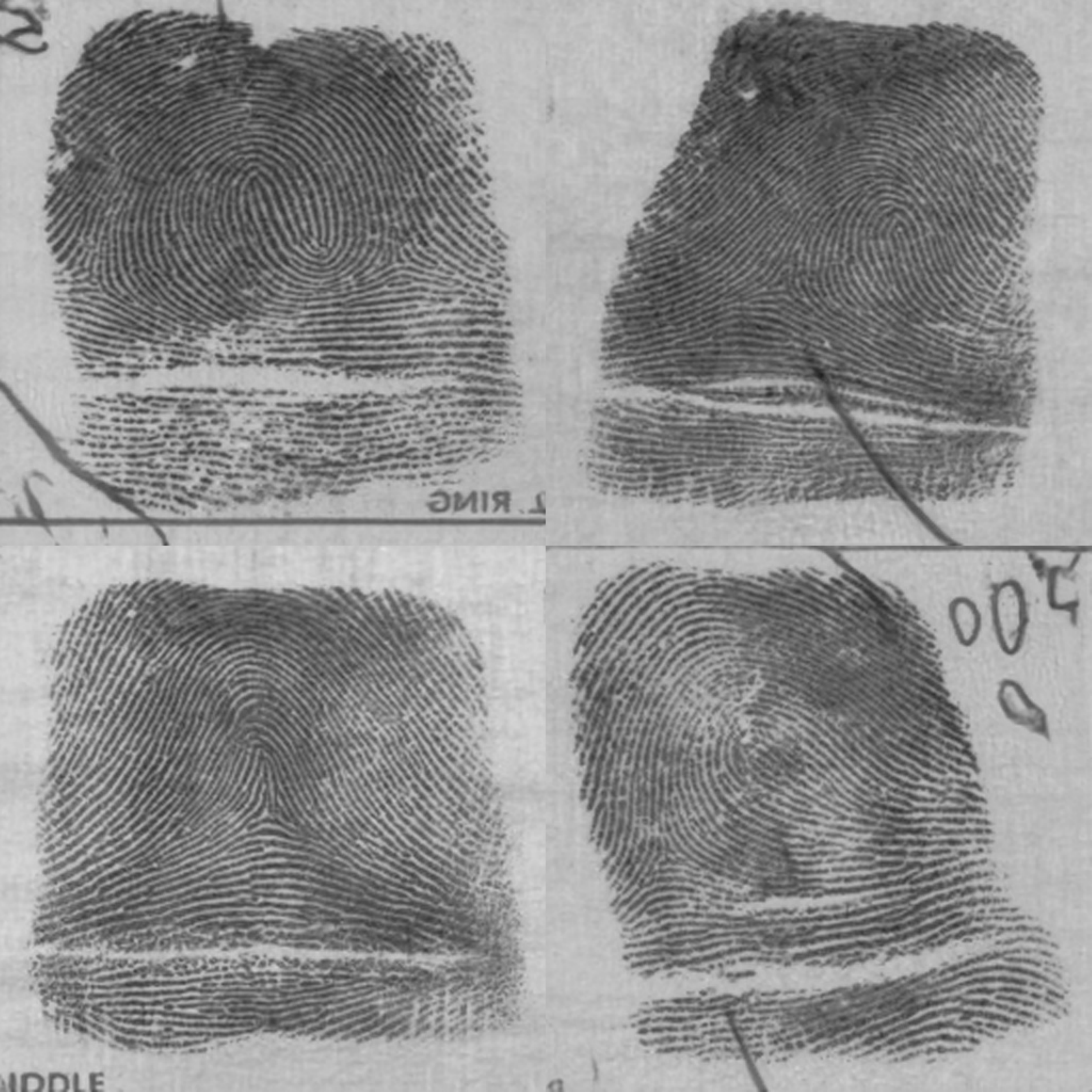}} \newline
\subfigure[Original]{\includegraphics[width=0.25\columnwidth,height=0.25%
\columnwidth]{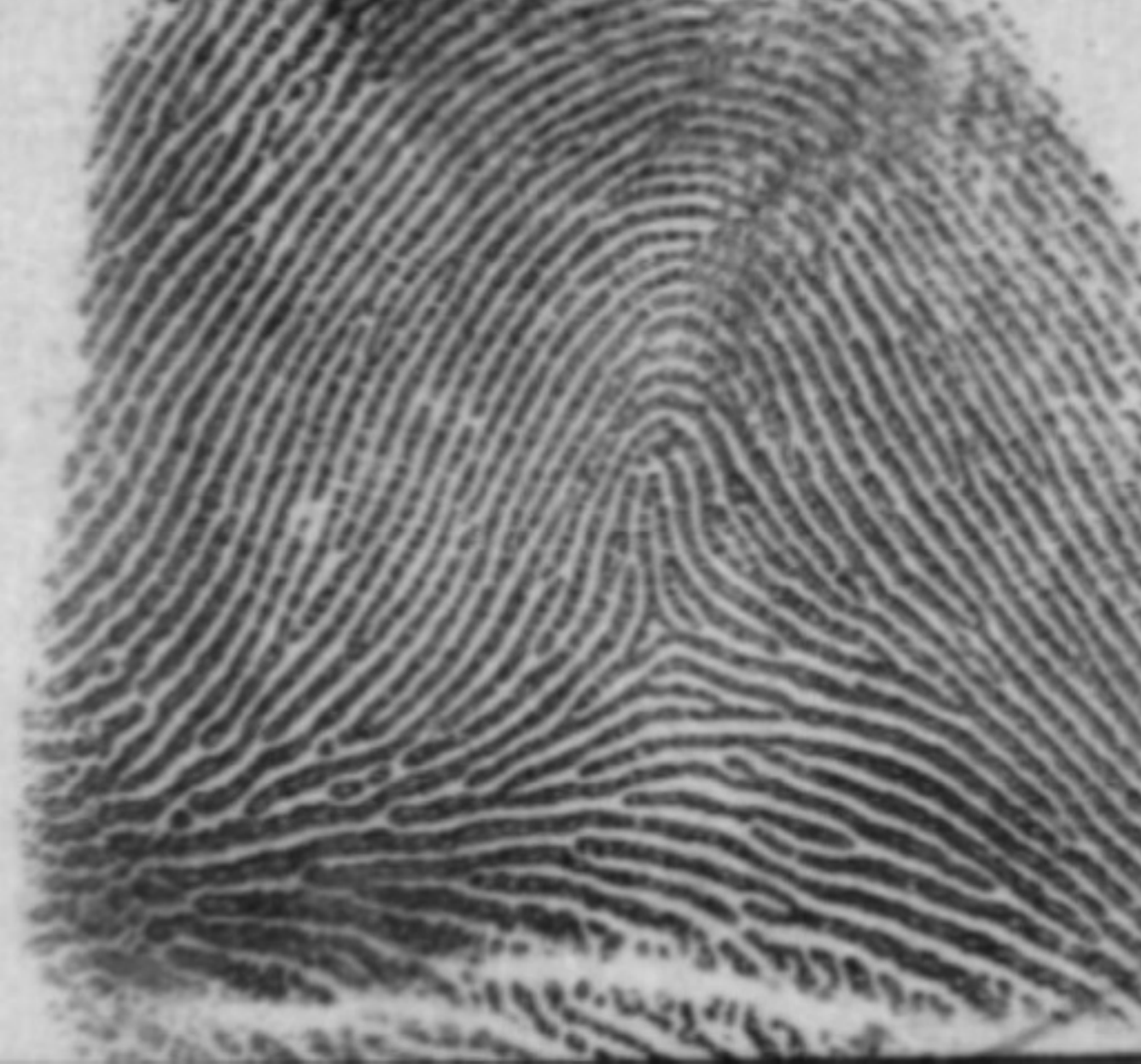}}\subfigure[Reconstructed]{%
\includegraphics[width=0.25\columnwidth,height=0.25%
\columnwidth]{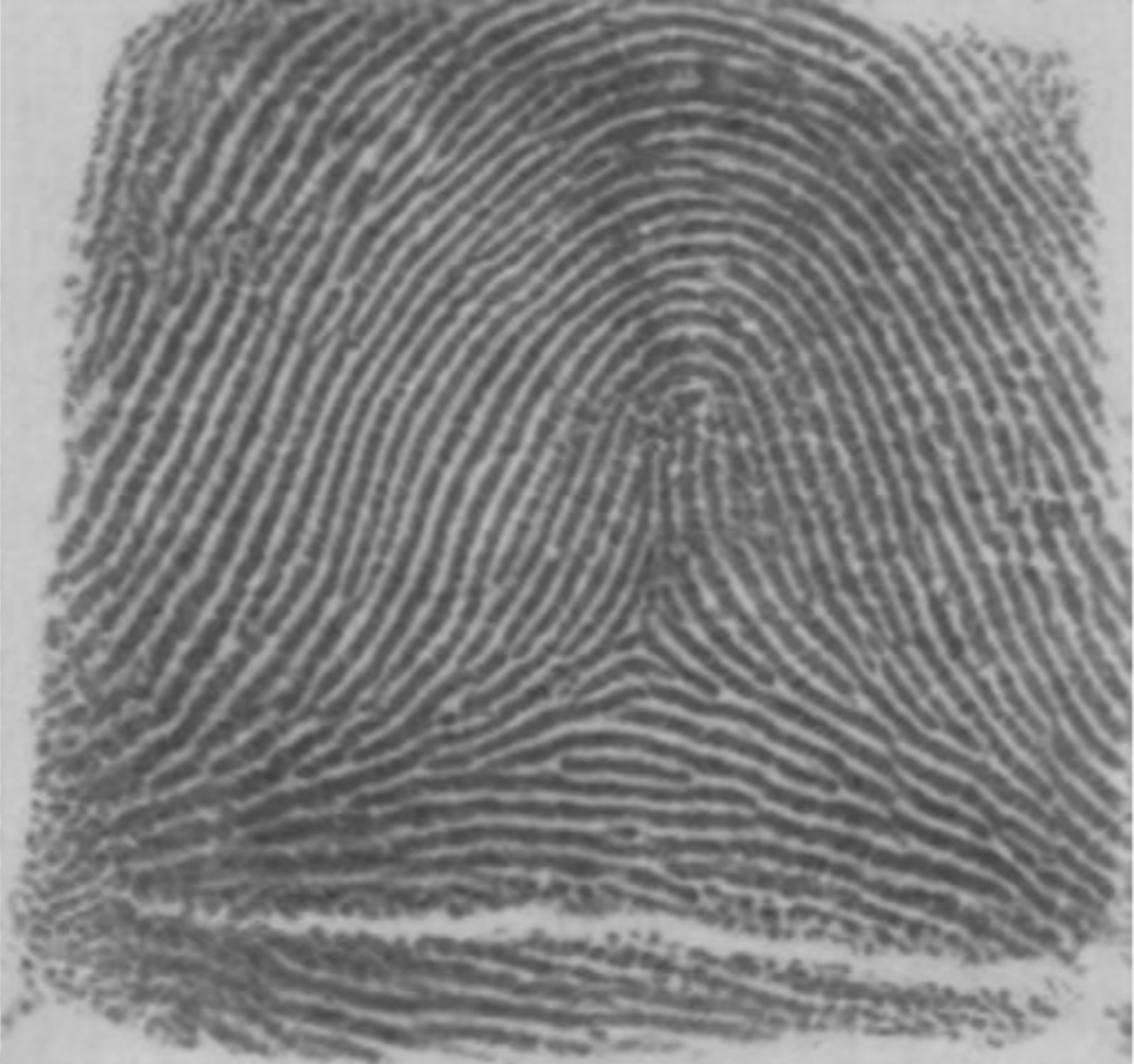}}\subfigure[Synthesized]{%
\includegraphics[width=0.25\columnwidth,height=0.25%
\columnwidth]{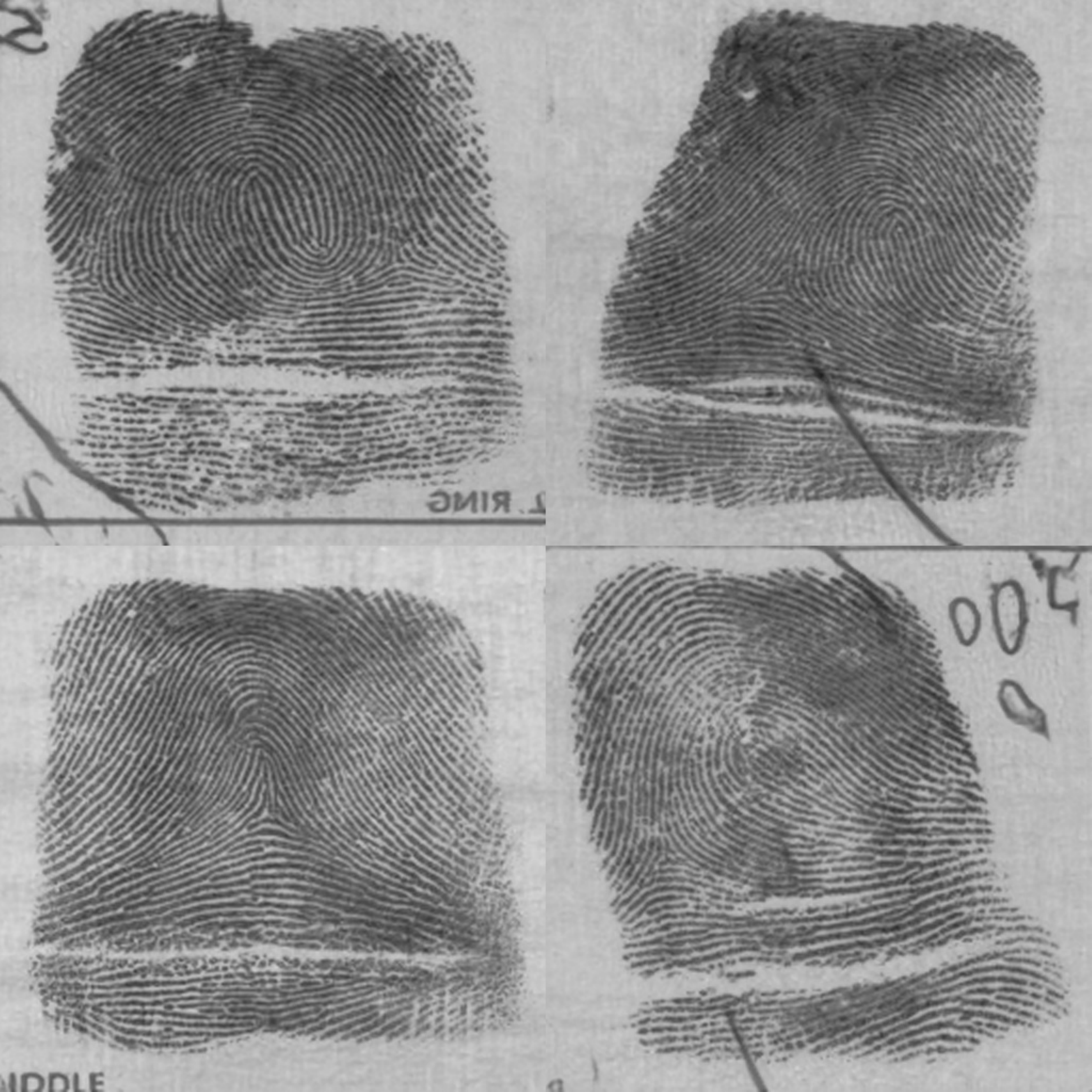}}\subfigure[Synthesized]{%
\includegraphics[width=0.25\columnwidth,height=0.25%
\columnwidth]{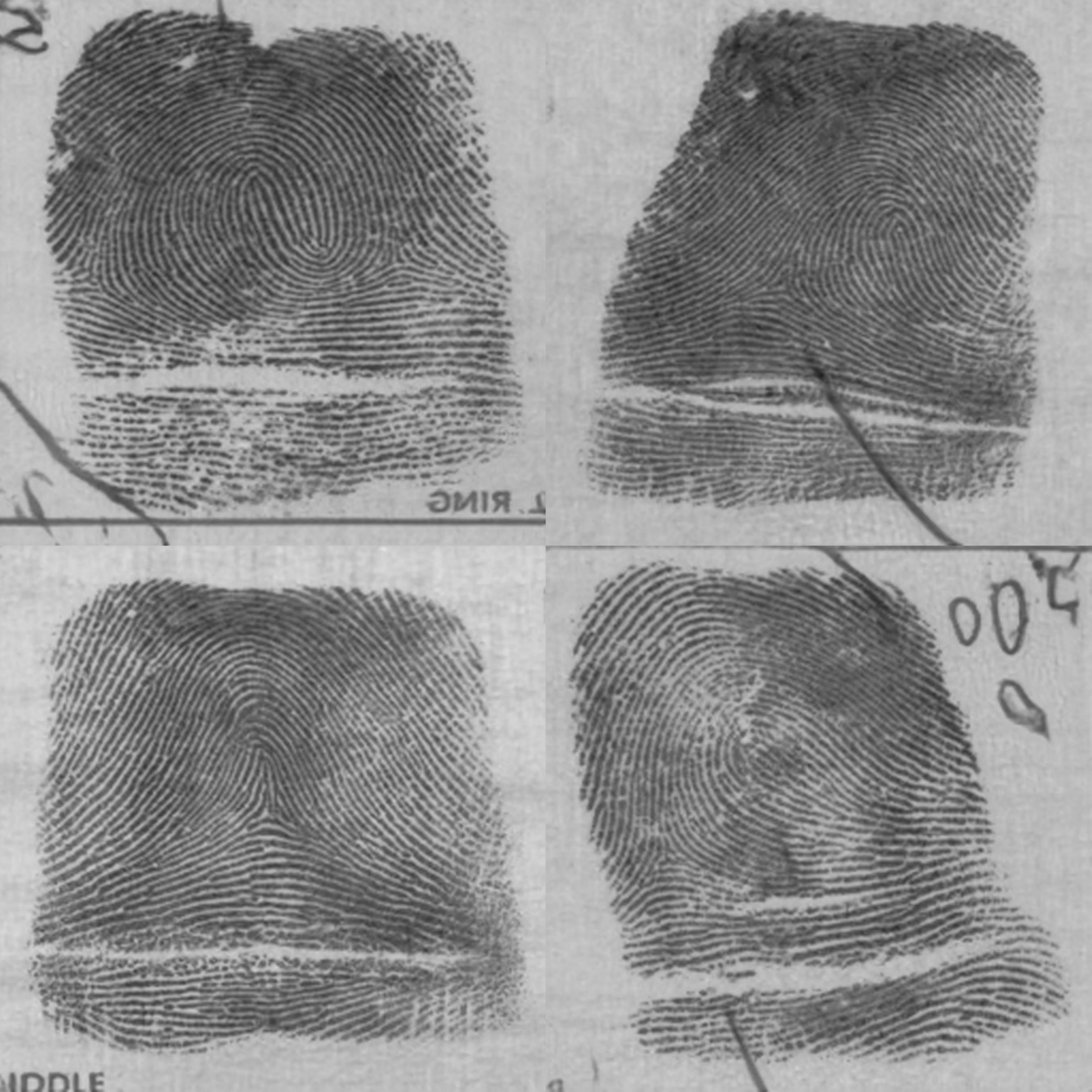}}
\caption{The proposed reconstruction and synthesis of fingerprints. (b) and
(f) are the reconstructions of (a) and (e), respectively, based on their
minutiae points. (a) and (e) were taken from the NIST SD4 dataset
\protect\cite{watson1992nist}. (c), (d), (g) and (h) are examples of
randomly synthesized fingerprints.}
\label{fig:teaser_image}
\end{figure}

To alleviate both cost and privacy issues, several approaches have been
proposed to create synthetic fingerprint datasets. Their goal is to
produce large datasets of synthetic fingerprints that can be used instead of
datasets of real fingerprints. The two core tasks associated with the generation of synthetic fingerprints are the synthesis \cite{cappelli2004sfinge,
zhao2012fingerprint, minaee2018finger, mistry2020fingerprint} and
reconstruction \cite{cappelli2007fingerprint,
feng2010fingerprint,cao2014learning, moon2021restore} that are shown in Fig. %
\ref{fig:teaser_image}. The goal of fingerprint synthesis is to generate
synthetic fingerprints that are as realistic as possible, while the goal of
fingerprint reconstruction is to generate synthetic fingerprints resembling
the source fingerprints as close as possible. In fingerprint reconstruction,
a set of features extracted from a given fingerprint, typically minutiae
\cite{cappelli2007fingerprint}, are provided, while in fingerprint synthesis
no input is given.

The common approach to fingerprint synthesis and fingerprint reconstruction is
model-based, where multiple computational models are used to estimate the
different attributes of the generated fingerprint: pattern area, orientation
image, and ridge pattern. Rendering is then applied to improve the realism
of the reconstructed fingerprint \cite%
{cappelli2004sfinge,cappelli2002synthetic,cappelli2007fingerprint,zhao2012fingerprint}%
. The fingerprints generated using these approaches lack the inherent
constraints between the orientation field, ridge valley structure, and
minutiae patterns, leading to unrealistic fingerprint images, easily
distinguishable from real fingerprints \cite{gottschlich2014separating}.
More recently, learning-based approaches, based mainly on the Generative
Adversarial Networks (GAN) architecture \cite{goodfellow2016nips}, have been
applied to fingerprint synthesis \cite{minaee2018finger, riazi2020synfi,
mistry2020fingerprint, bahmani2021high}. Contrary to model-based approaches,
GANs learn to generate fingerprint images adhering to the probability
distribution of \textit{fingerprint images}, rather than the attributes of
the fingerprints, in the training set. Although such approaches improve the
realism of the generated fingerprint images, the synthesized fingerprints do
not adhere to the fundamental attributes of real fingerprints as the
synthesis is based on random inputs. The distributions of the minutiae of
the fingerprints are not random, and different types of fingerprints have
different distributions of the minutiae. Therefore, the distributions of
minutiae of synthetically generated fingerprints do not conform to such
distributions. Another limitation of both the synthesis and the fingerprint reconstruction
approaches is their inability to control the visual attributes of a
fingerprint, such as the presence or absence of fingerprint artifacts,
background noise, and fingerprint shape. By altering fingerprint
attributes, a user can generate large datasets of fingerprints to address
image-level and minutiae-level fingerprint matching, artifact removal,
minutiae extraction, and fingerprint enhancement.

In this work, we propose a joint framework for fingerprint synthesis and
reconstruction, allowing one to create a wide range of synthetic fingerprint
datasets for a gamut of applications. The core of our approach is a
fingerprint generator based on the Stylegan2 architecture \cite%
{karras2020analyzing}, which is trained on a large dataset of fingerprints.
The generator produces a realistic fingerprint image, based on a latent vector input. A random fingerprint can be synthesized by feeding the
generator with a random vector. For fingerprint reconstruction, we propose
an encoder network to encode minutiae attributes as latent vectors, such
that the fingerprint generator can reconstruct the source fingerprint. To
preserve the identity of the synthesized and reconstructed fingerprints, we
derive a novel fingerprints-oriented perceptual loss based on the FingerNet
CNN \cite{tang2017fingernet}. We also present a novel approach for modifying
particular visual attributes of the fingerprint, such as blobs and dry skin
artifacts, while preserving its identity. By applying this framework, we
synthesize and share the SynFing dataset, consisting of 100K pairs of
fingerprint image, each pair having the same identity but different visual
attributes. This dataset can be used to train, validate, and spoof COTS
fingerprints verification systems.

In particular, we propose the following contributions:

\begin{itemize}
\item We present a novel joint framework for fingerprints synthesis and
reconstruction based on Generative Adversarial Networks and the encoding of
fingerprint attributes.

\item The proposed encoder and generator are trained using a novel minutiae
encoding and fingerprint-oriented perceptual loss based on FingerNet CNN
\cite{tang2017fingernet}, which has been shown to improve the accuracy of
reconstruction.

\item The framework introduces a novel computational approach to modify and
manipulate the visual attributes of the generated fingerprint image. This
allows to generate multiple fingerprint images for each fingerprint identity.

\item We introduce and share the large-scale SynFing dataset of synthetic
fingerprints that can be applied to fingerprint matching, artifacts removal,
and fingerprint spoof detection.

\item The proposed framework was experimentally shown to outperform
contemporary approaches for fingerprint synthesis and fingerprint reconstruction.
\end{itemize}

\section{Related work}

\label{sec:Previous and Related work}

\subsection{Fingerprint Synthesis}

Cappelli et al. \cite{cappelli2000synthetic} were the first to propose a
model-based approach to fingerprint synthesis, using multiple generation
steps, generating directional and density maps, ridge patterns, adding
noise, and rendering. They use the fingerprint's type, size, and singular
points as input. A Gabor-like space-variant filter and a modified Zero Pole
model were applied to generate a near-binary fingerprint image. Then,
task-specific noise is added to generate a realistic gray-scale
representation of the fingerprint. Although these methods are relatively
simple and do not require a large fingerprint database for model
calibration, they generated unrealistic fingerprints. In particular,
Gottschlich et al. \cite{gottschlich2014separating} demonstrated their
ability to differentiate between genuine and synthetic fingerprints by
examining the distribution of the minutiae points.

Recent learning-based approaches for fingerprint synthesis utilized
generative adversarial networks (GANs), where the generation is based on
processing a noise vector through a generative model to generate a synthetic
fingerprint image. The generative model is trained using a large fingerprint
dataset. Minaee et al. \cite{minaee2018finger} used the DCGAN \cite%
{radford2015unsupervised} architecture with a total variation regularization
term to impose connectivity within the generated images, while Mistry et al.
\cite{mistry2020fingerprint} applied an Improved Wasserstein GAN (IWGAN)
\cite{gulrajani2017improved}. Recently, Bahmani et al. \cite{bahmani2021high}
introduced the Clarkson Fingerprint Generator based on the StyleGan \cite%
{karras2019style} architecture.

\subsection{Fingerprint Reconstruction}

Fingerprint reconstruction is the synthesis of fingerprints based on given
attributes, such as minutiae templates. The reconstructed fingerprint aims
to reproduce the minute locations and distribution in the original
fingerprint. Contemporary reconstruction schemes follow two steps: first,
reconstruct the orientation field based on minutiae, and then reconstruct
the ridge pattern using the reconstructed orientation field. Cappelli et al.
\cite{cappelli2007fingerprint} proposed to reconstruct the grayscale image
directly using the minutiae. The orientation field was reconstructed by a
zero-pole model, followed by iterative Gabor filtering of the minutiae image
initialized by the local minutiae pattern. Using a disk-shaped structuring
element, Feng et al. \cite{feng2010fingerprint} predicted the local
orientation values, where the rigid pattern was reconstructed using the
amplitude and frequency modulated model (AM-FM) \cite{larkin2007coherent}.
Cao et al. \cite{cao2014learning} proposed a dictionary-based approach to
fingerprint reconstruction, where a dictionary of orientation patches is
used to reconstruct the orientation field from minutiae, while the
continuous phase patch dictionary is used to reconstruct the ridge pattern.
Similarly to fingerprint synthesis, recent fingerprint restoration models
utilize the GANs generation approach. Kin et al. \cite{kim2018reconstruction}
applied conditional GANs to minutiae images by formulating fingerprint
reconstruction as an image-to-image translation, in which a fingerprint
image is generated from an image containing minutiae information. Similarly,
Moon et al. \cite{moon2021restore} use Pix2Pix \cite{isola2017image} to
improve the reconstruction accuracy. However, the matching accuracy of the
reconstructed fingerprints was significantly inferior to that of the original
fingerprint images. Moreover, none of these works made their reconstruction
models publicly available, hindering the reproducibility and the fair
comparison between models, as their results are reported for private
datasets only \cite{kim2018reconstruction, moon2021restore}.
\begin{figure*}[th]
\centering\includegraphics[width=\textwidth]{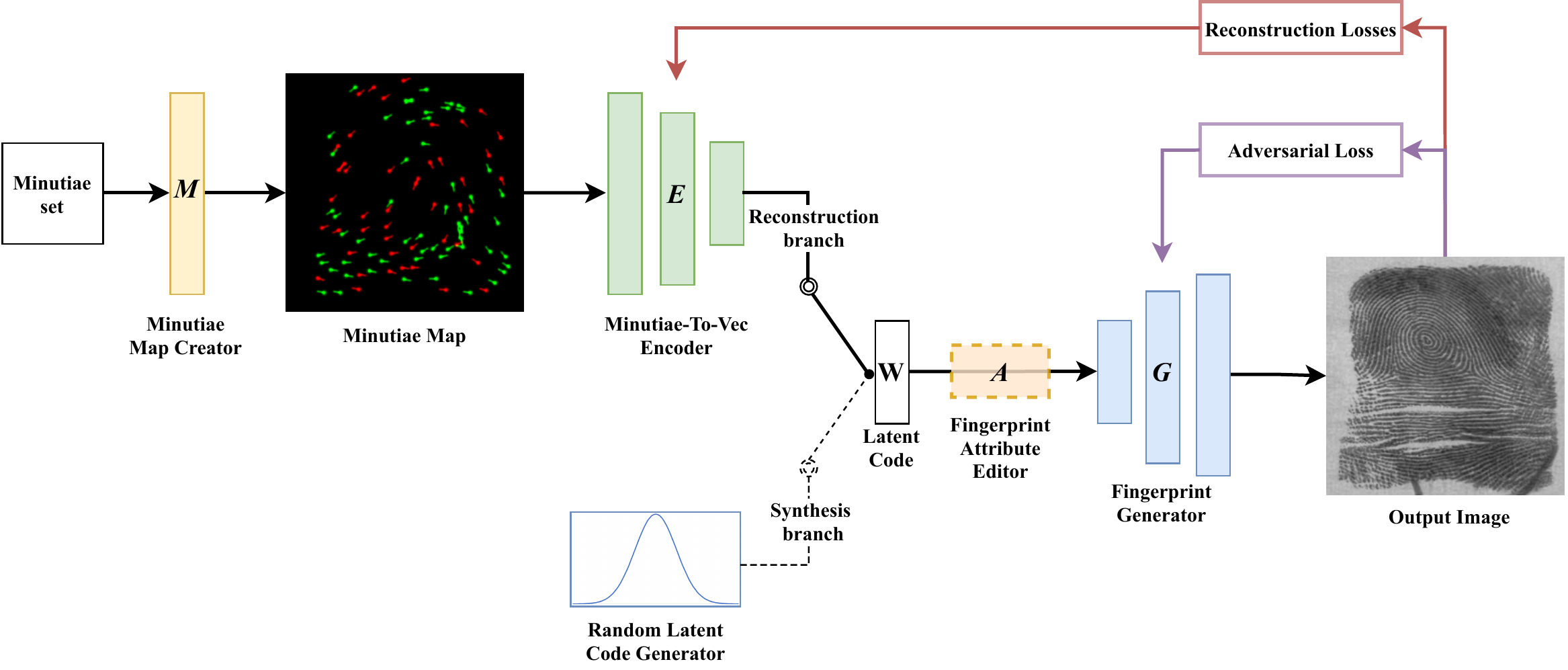}
\caption{\textbf{The proposed framework.} The proposed fingerprint generator
backbone $\mathbf{G}$ is based on the Stylegan2 architecture. For
fingerprint synthesis, the generator is given a normally distributed latent
vector $\mathbf{w}$. For reconstruction, we first encode the minutiae
attributes as a latent vector $\mathbf{w}$ using the Minutiae-To-Vec encoder
network $\mathbf{E}$. A fingerprint attribute editor $\mathbf{A}$ is added
before the generator's input layer, allowing to modify particular attributes
of the generated fingerprints, by modifying the latent vector $\mathbf{w}$.}
\label{fig:framework}
\end{figure*}

\section{Fingerprint Synthesis and Reconstruction}

\label{sec:Fingerprint Synthesis and Reconstruction}

We propose a joint framework for fingerprint synthesis and reconstruction,
whose overview is shown in Fig.~\ref{fig:framework}. Our fingerprint
generator, detailed in Section \ref{subsec:Fingerprint_Generator}, is based
on the StyleGAN2 architecture \cite{karras2020analyzing} ($\mathbf{G}$ in
Fig.~\ref{fig:framework}), trained using the NIST SD14 fingerprint dataset
to synthesize fingerprint images. To generate a new fingerprint image, we
feed the generator $\mathbf{G}$ with a latent vector $\mathbf{w}$. In the
synthesis branch, a normal distribution generator is used to randomly
generate $\mathbf{w}$, which is input into the generator $\mathbf{G}$ to
synthesize a realistic fingerprint image. In the reconstruction branch,
detailed in Section \ref{subsec:Reconstruction}, we train a Minutiae-To-Vec
encoder network $\mathbf{E}$ to encode the minutiae information extracted
from a particular fingerprint into a latent vector $\mathbf{w}$, used to
reconstruct the original fingerprint. The proposed Fingerprint Attribute
Editor $\mathbf{A}$ (Section \ref{subsec:AttributesModification}) is added
before the input layer of the generator $\mathbf{G}$, to allow the user to
manipulate the latent code $\mathbf{w}$ to modify particular attributes of
the generated fingerprint, while preserving its identity.

\subsection{Fingerprint Generator}

\label{subsec:Fingerprint_Generator}

The proposed fingerprint generator is based on the StyleGAN2 architecture
\cite{karras2020analyzing} that is given a latent vector $z\in\mathbb{R}
^{512}\sim P_{z}$, where $P_{z}$ is a multivariate Gaussian distribution.
Unlike traditional generators, in which $\boldsymbol{z}$ is directly passed
into the convolutional upscaling layers, the StyleGAN2 generator uses a
multistage input block to map $\boldsymbol{z}$ to a style vector $%
\boldsymbol{y}$. Thus, the latent vector $\boldsymbol{z}$ is mapped to an
intermediate latent vector $\boldsymbol{w}$, using an 8-layer MLP network $f:%
\boldsymbol{z}\rightarrow \ \mathbf{w}$, then learned affine transformations
specialize $\mathbf{w}$ to multiple style vectors $\boldsymbol{y}$. Each
style vector is fed into a different convolution layer of the generator. The
output of the generator is a grayscale image $\mathbf{I\in\mathbb{R}}^{h\times w}$.

\subsection{Fingerprint Reconstruction}
\label{subsec:Reconstruction}
Let $\boldsymbol{x}\in\mathbb{R}^{h\times w\times 3}$ be a fingerprint image and $S_{x}=\left\{ {\theta
_{ij},T_{ij}}\right\} $ is the set of minutiae extracted from $\boldsymbol{x}
$ at the points $\left\{ i,j\right\} $, where $\theta _{ij}$ and $T_{ij}$
are the direction and class of the minutia, respectively. The reconstruction model
reconstructs the fingerprint image $\boldsymbol{x}$ given $S_{x}$. We cast
the fingerprint reconstruction as image-to-image translation \cite%
{richardson2021encoding, luo2020time} where the minutiae set is first
converted to a minutiae map $\mathbf{M}_{\mathbf{x}}\mathbf{\in\mathbb{R}}^{h\times w\times 3}$ following \cite{kim2018reconstruction, ma2017pose,
tang2019cycle}. $\mathbf{M}_{\mathbf{x}}$ is encoded by a latent vector $%
\boldsymbol{w}\mathbf{\in\mathbb{R}}^{512}$ using a Minutiae-To-Style CNN (Section \ref%
{subsubsec:Minutiae-To-Vec}), that is fed to the pretrained generator $%
\mathbf{G}$ to reconstruct the input image $\boldsymbol{x}$.

\subsection{Minutiae-To-Vec Encoder}

\label{subsubsec:Minutiae-To-Vec}

Given a minutia set $S_{x}$ of the fingerprint image $\boldsymbol{x}\in\mathbb{R}^{h\times w\times 3}$, its corresponding minutia map $\mathbf{M}_{\mathbf{x}}\in\mathbb{R}
^{h\times w\times 3}$ is given by%
\begin{equation}
\mathbf{M}_{\mathbf{x}}^{i,j}=\left\{
\begin{tabular}{ll}
$\left\vert l(\theta _{ij}),0,0\right\vert $ & \text{$T_{ij}$ is a
Bifurcation }point \\
$0,l(\theta _{ij}),0$ & \text{$T_{ij}$ is a Termination }point \\
$0,0,l(\theta _{ij})$ & \text{$T_{ij}$ is a Singular point} \\
$0,0,0$ & \text{otherwise}%
\end{tabular}%
\ \ \ \ \ \right. .  \label{eqn:M_RGB}
\end{equation}

For each location $(i,j)\in \ S_{x}$ we draw a line $l(\theta _{ij})$ in
orientation $\theta _{ij}$ that is drawn at the point in the channel
corresponding to the class of minutiae $T_{ij}$ as in Eq. \ref{eqn:M_RGB}. $%
\mathbf{M}_{\mathbf{x}}$ is convolved with a Gaussian kernel $G(\sigma )$ to
create a smoother minutiae map. Minutia maps are shown in Fig. \ref%
{fig:minutiae_combinde}. The encoder $\mathbf{E}$ encodes $\mathbf{M}_{%
\mathbf{x}}$ as the latent vector $\boldsymbol{w}\mathbf{\in\mathbb{R}}^{512}$, used by the generator to generate a fingerprint with the same
attributes as the original. These properties include the finger type,
fingerprint shape, internal interactions between minutiae points, singular
areas, and the locations, types and orientations of the minutiae points. We
applied the ResNet50 CNN \cite{he2016deep} as the encoder network $\mathbf{E}
$ to compute $\boldsymbol{w,}$ by adding a fully-connected layer following
its last convolutional layer.
\begin{figure}[th]
\centering%
\subfigure[Fingerprint
image]{\includegraphics[width=0.33\columnwidth,height=0.33\columnwidth]{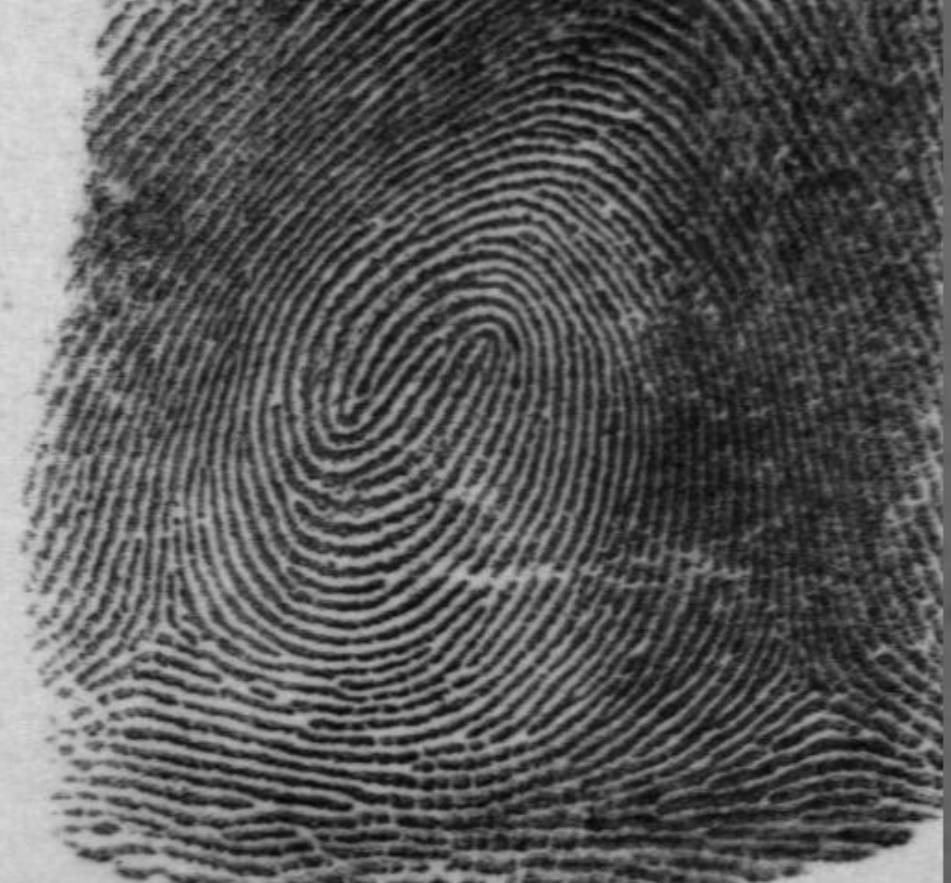}}%
\subfigure[Minutiae
map]{\includegraphics[width=0.33\columnwidth,height=0.33\columnwidth]{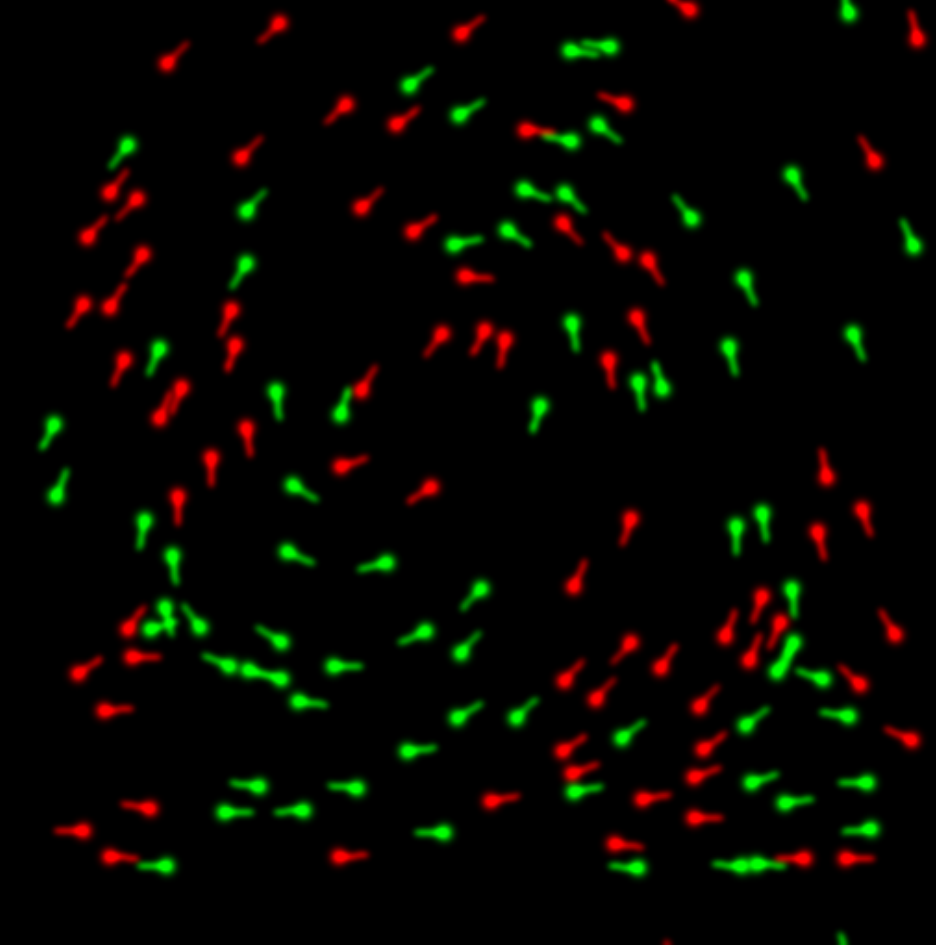}}%
\subfigure[Overlaid minutiae
map]{\includegraphics[width=0.33\columnwidth,height=0.33\columnwidth]{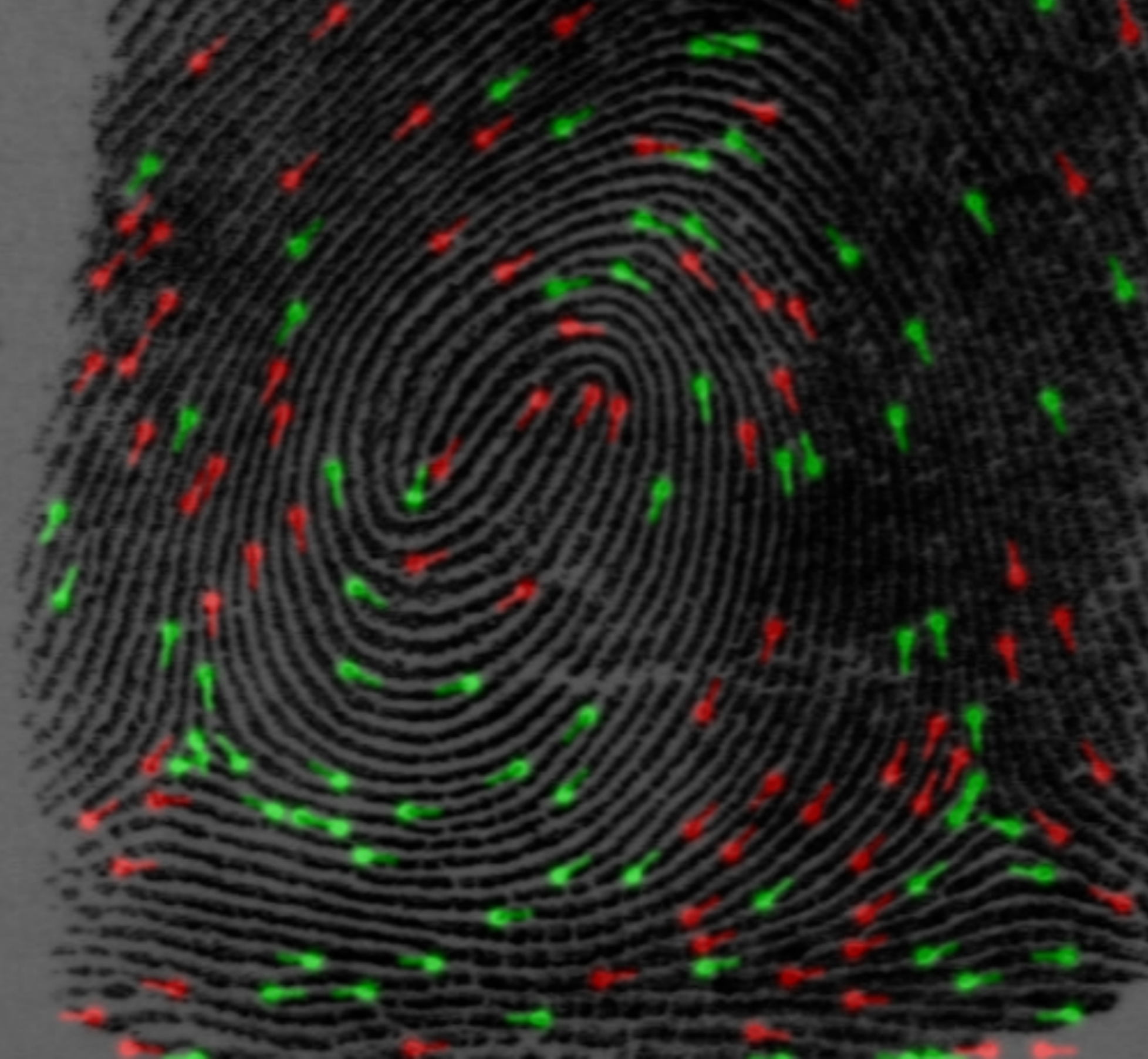}}
\vspace{-0em}%
\caption{A fingerprint image from the NIST SD4 dataset and its minutiae map $%
\mathbf{M}_{\mathbf{x}}$. The green and red points are the Termination and
Bifurcation minutiae points, respectively. The minutiae orientation is
encoded by the angle of the line drawn at each point.}
\label{fig:minutiae_combinde}
\end{figure}

\subsection{Fingerprint Attributes Editor}

\label{subsec:AttributesModification}

The proposed fingerprint generation approach allows modifying the attributes
of the generated fingerprints, while preserving their identity (minutiae,
rigid pattern, and finger type). Such attributes are the appearance of
artifacts, acquisition position, and appearance of noises in the background.
Thus, a user can generate multiple \textit{different} impressions of the
same fingerprint, remove noise and artifacts, and apply image-level
fingerprint matching.

For that, we follow the SeFa approach by Shen and Zhou \cite%
{shen2021closedform}, which is an unsupervised closed-form method to
identify semantic patterns in the latent space $\mathbf{w}$. It is used to
estimate the latent semantic directions in $\mathbf{w}$ that modify
particular attributes of the fingerprint while preserving their identity.
For instance, by varying the first dimension of $\mathbf{w}$, we can control
the appearance of scribbles in the fingerprint background, while varying the
second dimension adds blobs and dry skin artifacts to the fingerprint, as
shown in Fig~\ref{fig:Sefa_examples}. To the best of our knowledge, ours is
the first study to introduce such capabilities to fingerprint generation.

\subsection{Losses and Training}

\label{subsec:training}

The proposed scheme is trained in two steps:

\textbf{Generator model training.} The first step is to train the ST2-based
generator model \cite{karras2020analyzing} detailed in Section \ref%
{subsec:Fingerprint_Generator} to synthesize random fingerprints. The
generator is trained from scratch using an uninitialized ST2 model. Training follows the GAN training scheme with an adversarial discriminator
loss. The discriminator is alternately fed real and generated fingerprint
images, which are used to update the weights of both the generator and
discriminator model. We use adaptive discriminator augmentation (ADA) \cite%
{karras2020training} to stabilize training.

\textbf{Minutiae-To-Vec Encoder.} The next step is to train the
Minutiae-To-Vec Encoder (Section \ref{subsubsec:Minutiae-To-Vec}) to embed
the minutiae as a latent vector. We freeze the fingerprint generator weights
to stabilize the training and use them to generate the reconstructed
fingerprint image. The encoder was trained using two losses: the pixelwise $%
L_{2}$ fingerprint reconstruction loss:

\begin{equation}
L_{2}(\mathbf{x})=\left\Vert \mathbf{x}-\mathbf{G}(\mathbf{E}(\mathbf{M}_{%
\mathbf{x}}))\right\Vert _{2},
\end{equation}%
where $\boldsymbol{x}$ is the original fingerprint image and $\mathbf{G}(%
\mathbf{E}(\mathbf{M}_{\mathbf{x}}))$ is the reconstructed fingerprint
image. $\mathbf{G}$ and $\mathbf{E}$\textbf{\ }are the generator and encoder
models, respectively. A common challenge in reconstruction is to preserve
the input's identity. For that, we apply a pretrained FingerNet CNN \cite%
{tang2017fingernet}%
\begin{equation}
L_{id}(\mathbf{x})=\left\vert F_{net}(\mathbf{x})-F_{net}(\mathbf{\tilde{x}}%
)\right\vert ,
\end{equation}%
where $F_{net}(\mathbf{x})$ is the output of the FingerNet CNN consisting of
minutiae detection scores, X and Y minutiae probabilities, minutiae
orientations, and the segmentation map. The overall reconstruction loss is
thus given by

\begin{equation}
L_{recon}(\mathbf{x})=L_{2}(\mathbf{x})+L_{id}(\mathbf{x}).  \label{equ:loss}
\end{equation}

\section{Experimental Results}

\label{sec:Experiments}

The proposed scheme was experimentally verified by applying it to benchmarks
and datasets of rolled fingerprint images used in contemporary
state-of-the-art synthesis and reconstruction schemes. The source-codes of
most of these schemes, such as Minaee at el. \cite{minaee2018finger}, Ross
\cite{ross2007template}, Li and Kot \cite{li2012improved} and Feng and Jain
\cite{feng2010fingerprint}, are unavailable, and their results are thus
cited from other publications.

\subsection{Datasets}

\label{subsec:Datasets}

\textbf{NIST SD14.} For fingerprint matching and classification purposes, we
used the NIST SD14 dataset \cite{watson2001nist}, which consists of
54,000 rolled fingerprint images obtained from 27,000 unique fingers (with
two impressions per finger). This dataset was used to train both the
Minutiae-To-Vec encoder and the proposed fingerprint generator.

\textbf{NIST SD4.} The NIST SD4 dataset \cite{watson1992nist} has been
commonly employed for testing and building automated fingerprint
classification systems. It consists of 4,000 rolled fingerprint images
obtained from 2,000 unique fingers (with two impressions per finger). Fingerprints are classified into five distinct types, each class containing
an equal number of prints (400). We utilized this dataset to assess our
framework's performance in both the synthesis and reconstruction tasks.

\textbf{SynFing.} The SynFing dataset consists of 100K pairs of synthetic
rolled fingerprints created using the proposed fingerprint generator and
attribute modifier. Each pair of impressions shares the same synthetic
identity but differs in visual attributes, such as scribbles and dry-skin
artifacts. We made the SynFing dataset available.

\textbf{CaoJain.} We came across only one open source work, the CaoJain
dataset, which includes 40,000 synthetic fingerprint images created by Cao
and Jain \cite{cao2018fingerprint}. We generated this dataset to compare and
contrast it with the proposed fingerprint synthesis approach.

\subsection{Implementation Details}

\label{sec:Implementation Details}

The proposed fingerprint generator was applied to the random input $%
\boldsymbol{w}\in\mathbb{R}^{512}$, generating an image $\boldsymbol{I}\in\mathbb{R}^{512\times 512}$. The NIST SD14 dataset was used to train the generator. It
was split to 20k, 5k, and 2k fingerprints for the train, validation and test
sets, respectively\textbf{.} We preprocessed both the NIST SD14 and NIST SD4
datasets by cropping the fingerprint regions using the FingerNet
segmentation CNN \cite{tang2017fingernet} and resized them to $\mathbb{R}
^{512\times 512}$. The set of minutiae was extracted using Verifinger SDK
11.1 \cite{VeriFinger}. In the creation of the Minutiae Map $\mathbf{M}_{%
\mathbf{x}}$ we encoded the bifurcation, termination, and singular points.
We drew a line 15 pixels long with an orientation determined by the minute
direction for every minute point. The resulting map was smoothed by a
Gaussian kernel with $\sigma =9$ pixels.

The generator network $\mathbf{G}$ was trained using the Adam optimizer ($%
\beta _{1},\beta _{2}=0,0.99$) with a batch size of 16. The learning rates
of the generator and discriminator were set to $0.0016$ and $0.0019$,
respectively. We used the ADA augmentation policy \cite{karras2020training}
with a probability of 0.6, and train the model for 480K epochs. The model
generates 45 fingerprint images per second on a single NVIDIA GTX 2080 TI
GPU. The input to the Minutiae-To-Vec encoder is the minutia map $\mathbf{M}%
_{\mathbf{x}}\in\mathbb{R}
^{512\times 512\times 3}$, and the output is a latent vector $\boldsymbol{w}%
\in\mathbb{R}
^{512}$. The encoder was trained using the same dataset (NIST SD14) and
splits as the generator. We use the Ranger optimizer with a batch size of 4
and a learning rate of $0.0001$ and train the model for 150K epochs.

\subsection{Fingerprint Synthesis}

\label{sec:Fingerprint Synthesis}
\begin{figure*}[t]
\begin{tabular}{lllllll}
\raisebox{+1.5\height}{\rotatebox[origin=b]{90}{NIST SD4}} & %
\includegraphics[width=0.28\columnwidth,height=0.28%
\columnwidth]{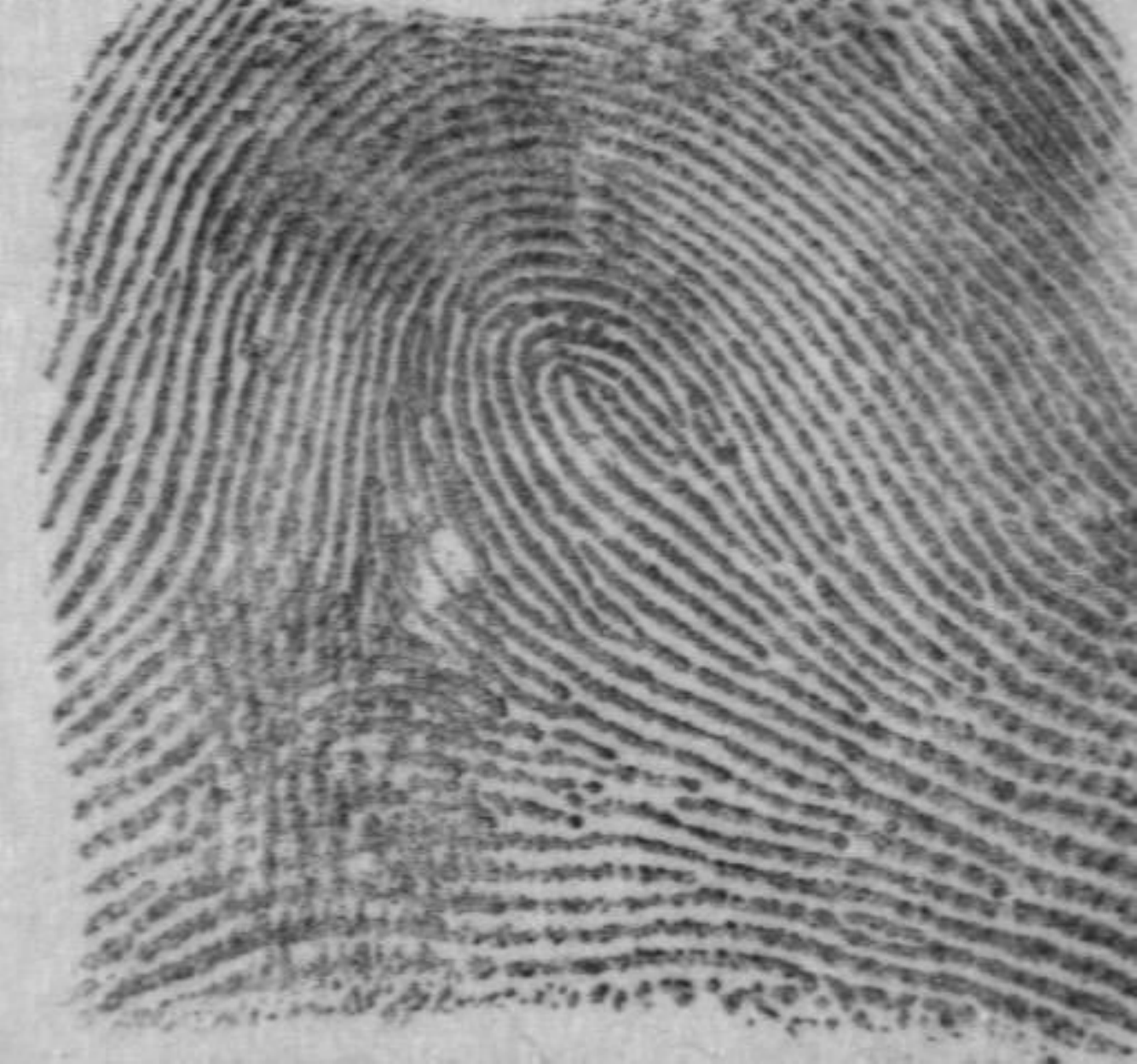} & %
\includegraphics[width=0.28\columnwidth,height=0.28%
\columnwidth]{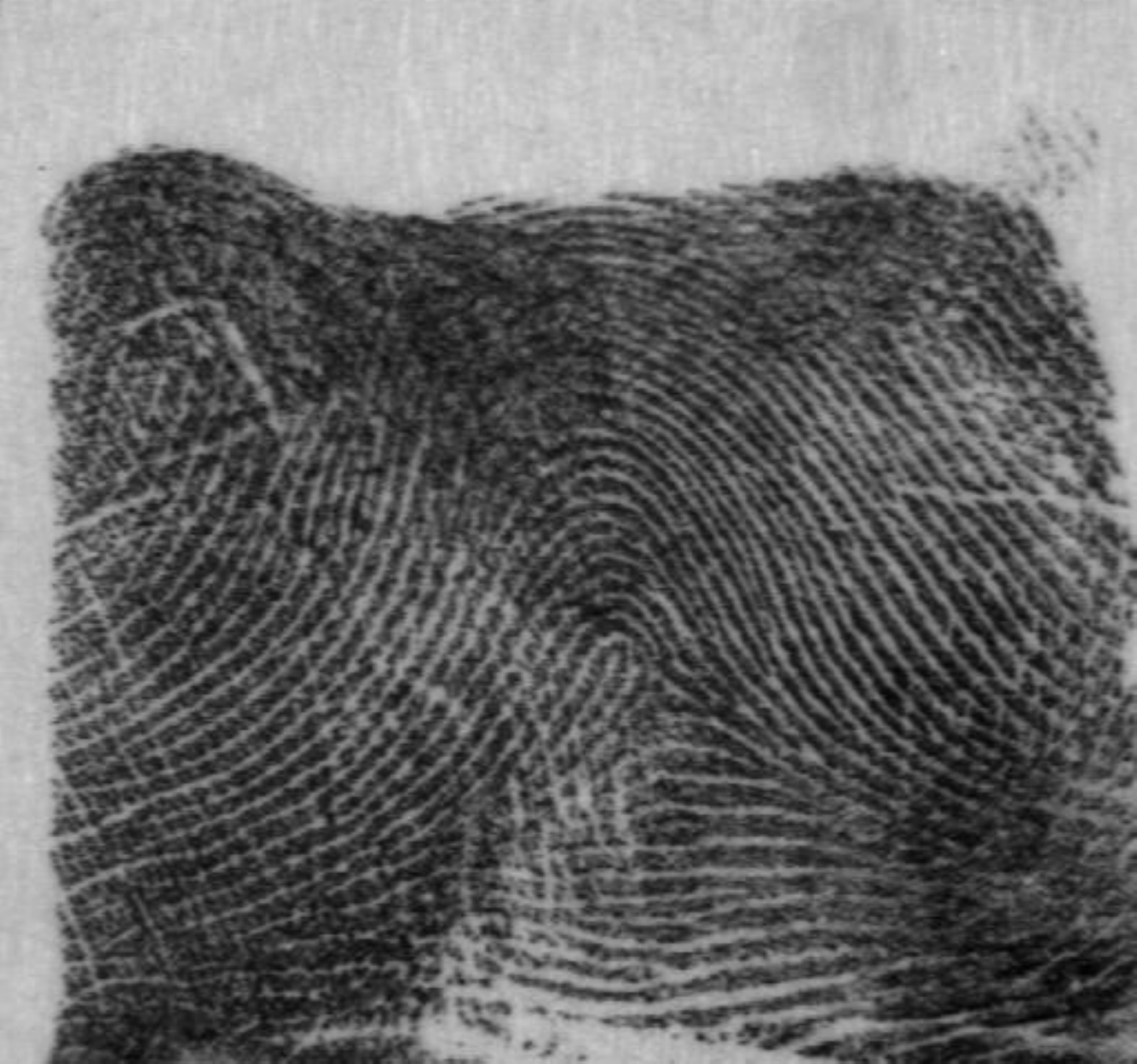} & %
\includegraphics[width=0.28\columnwidth,height=0.28%
\columnwidth]{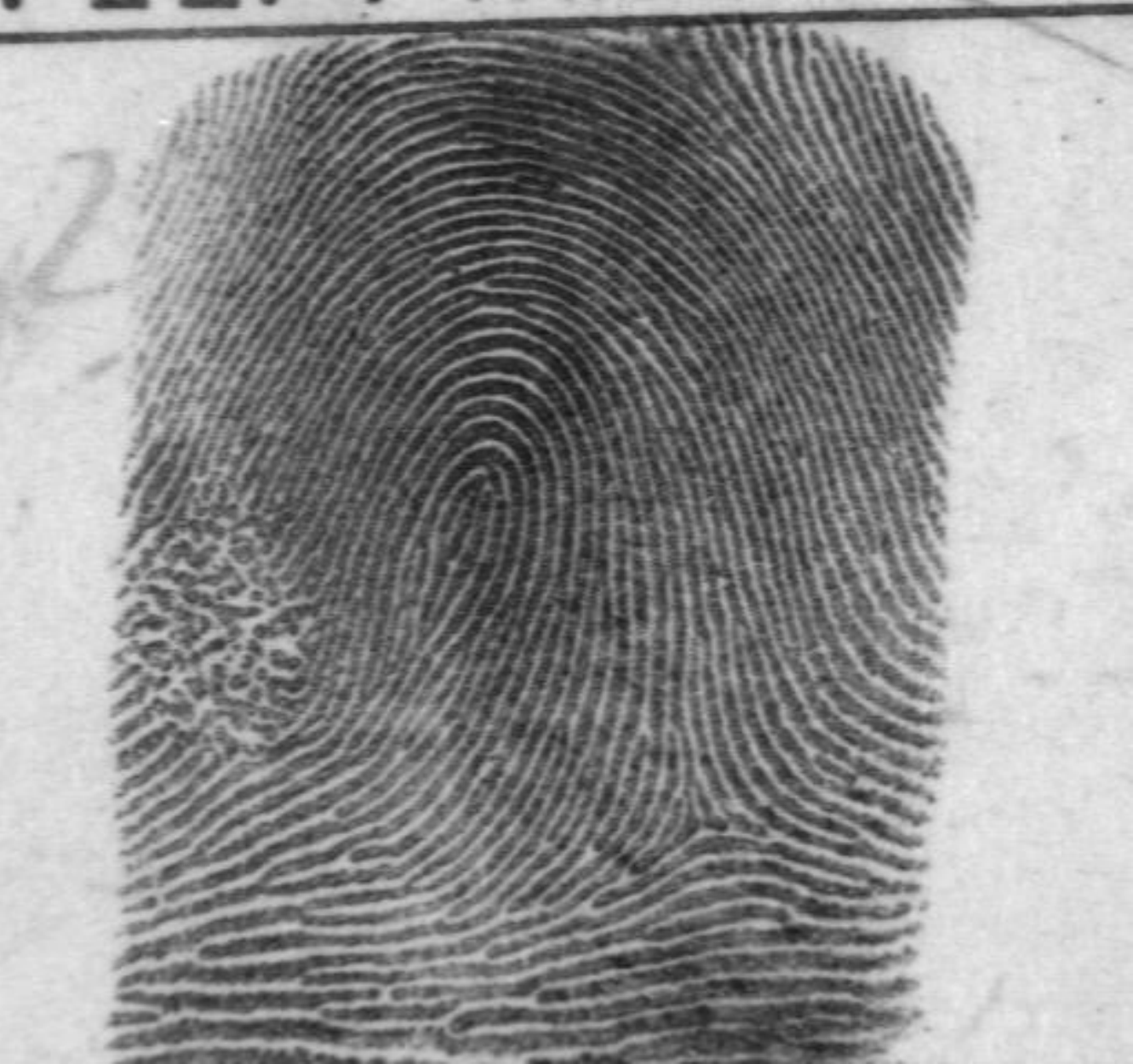} & %
\includegraphics[width=0.28\columnwidth,height=0.28%
\columnwidth]{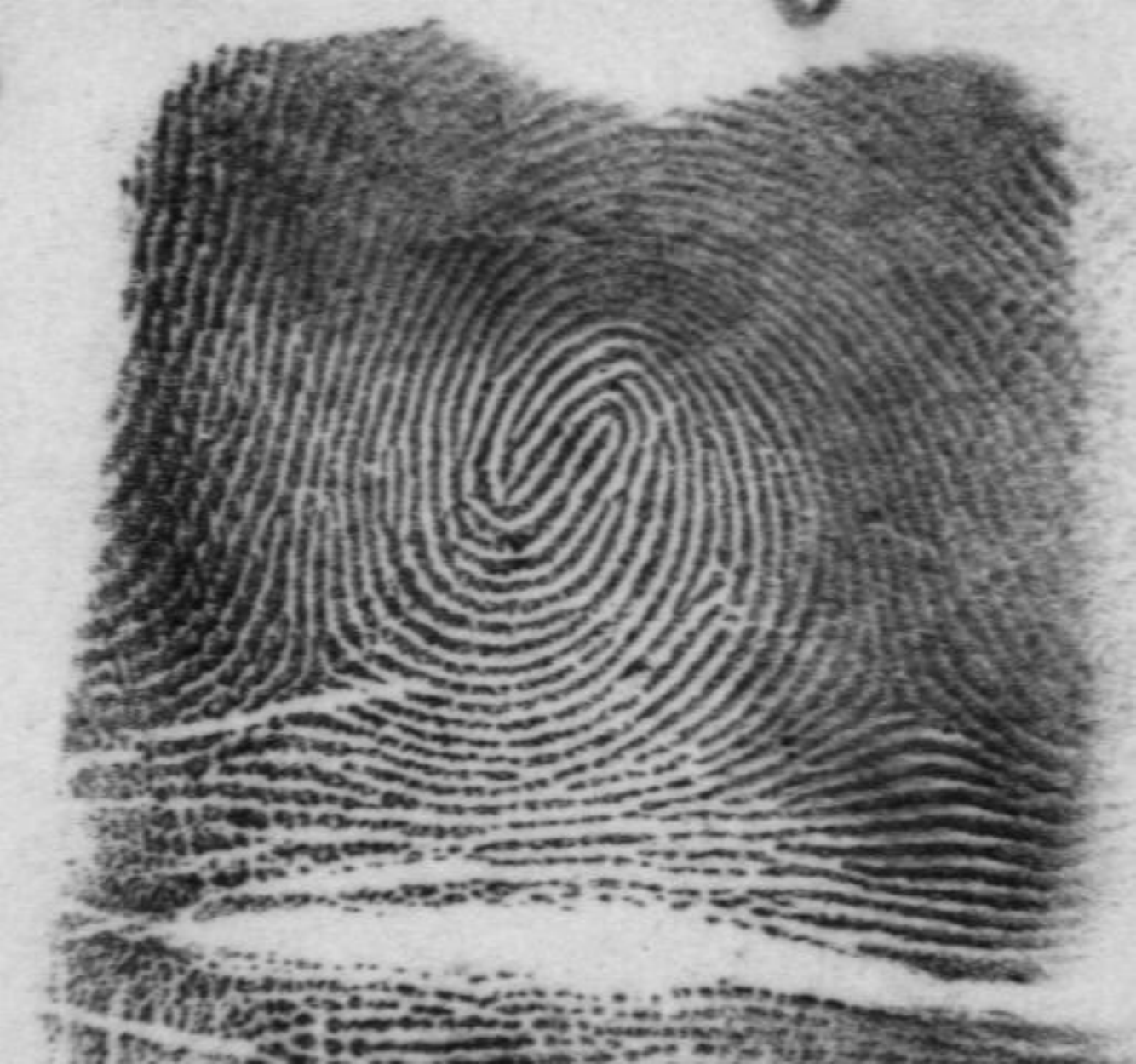} & %
\includegraphics[width=0.28\columnwidth,height=0.28%
\columnwidth]{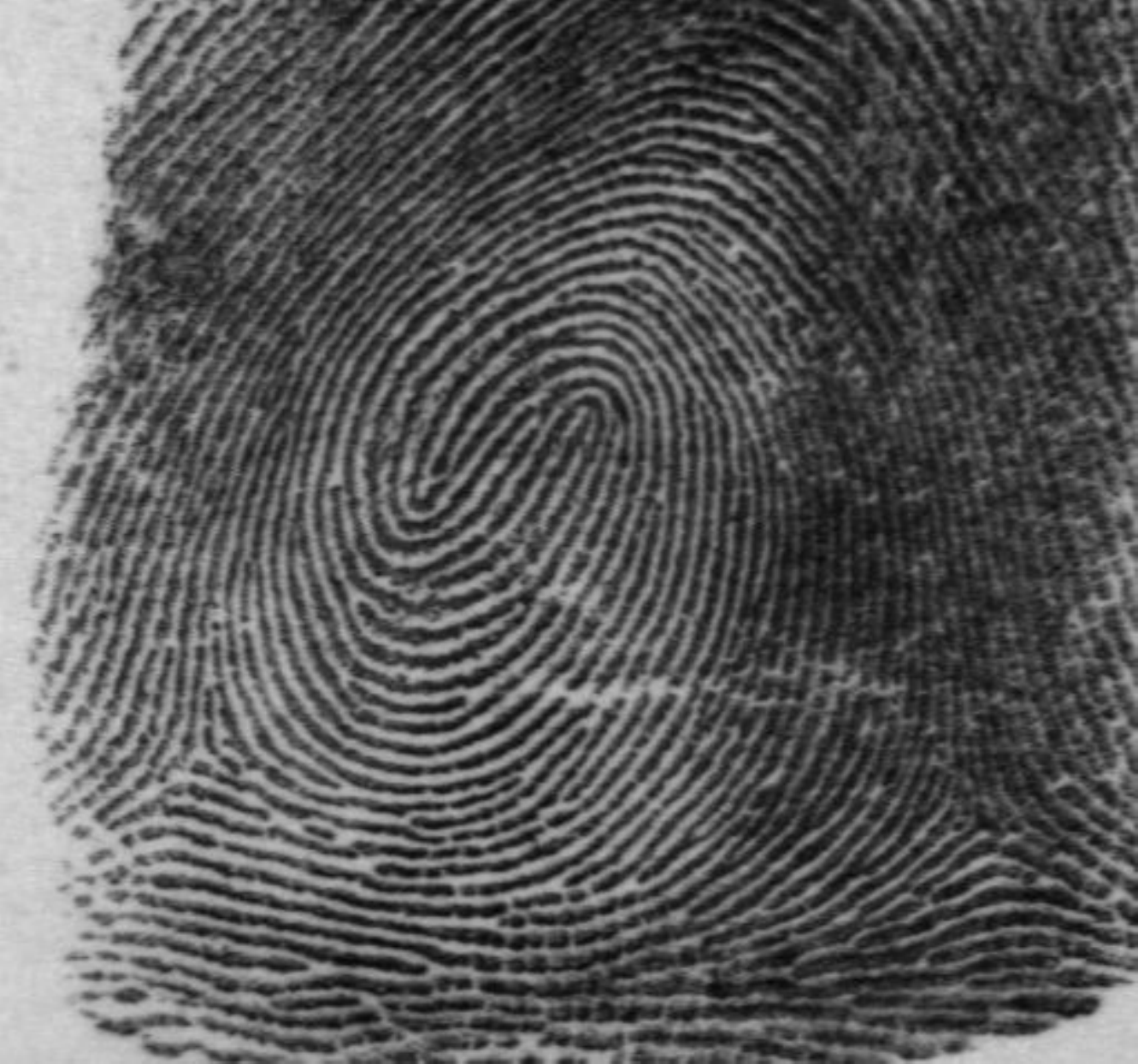} & %
\includegraphics[width=0.28\columnwidth,height=0.28%
\columnwidth]{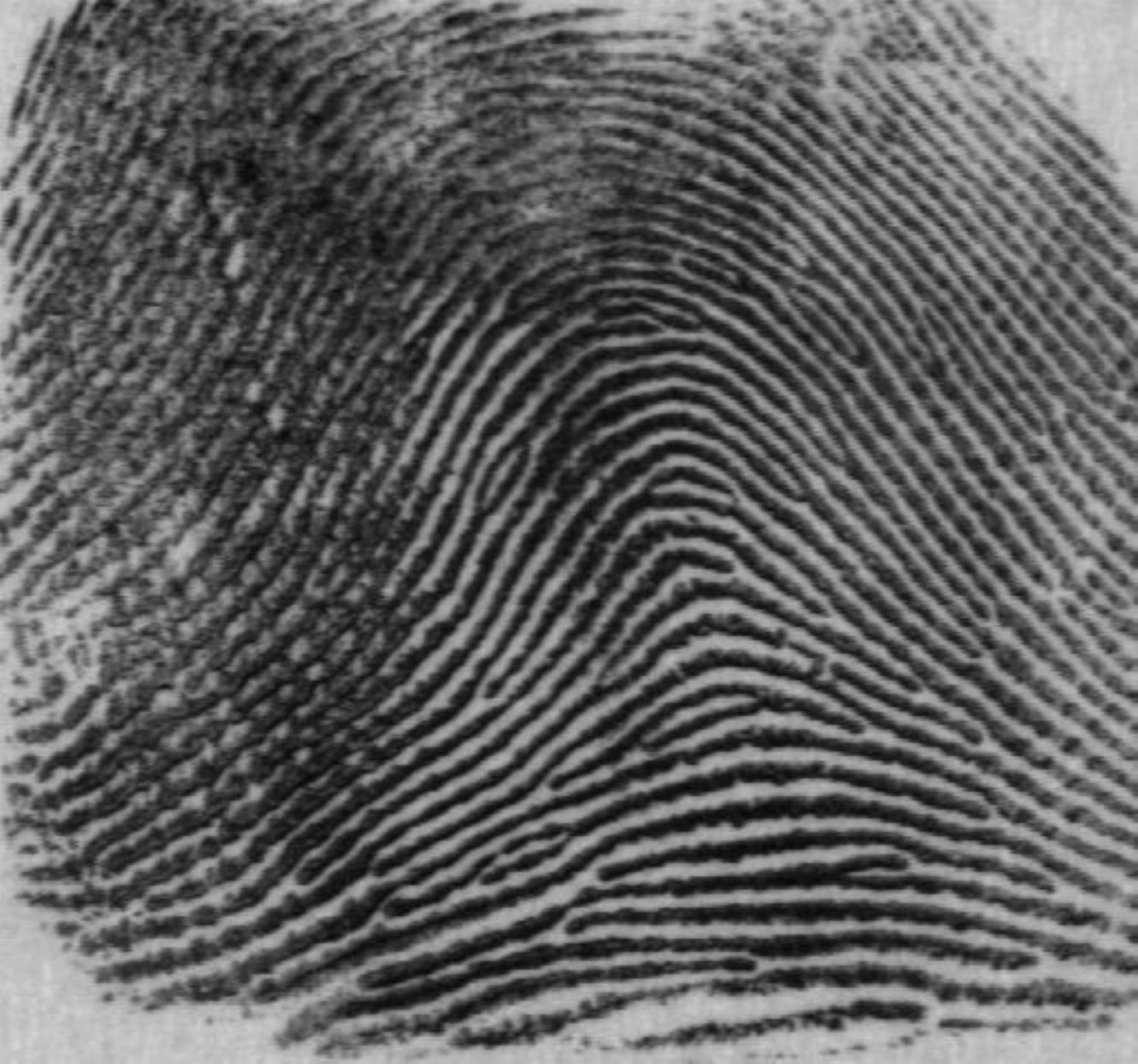} \\
\raisebox{+2.5\height}{\rotatebox[origin=b]{90}{SynFing}} & %
\includegraphics[width=0.28\columnwidth,height=0.28%
\columnwidth]{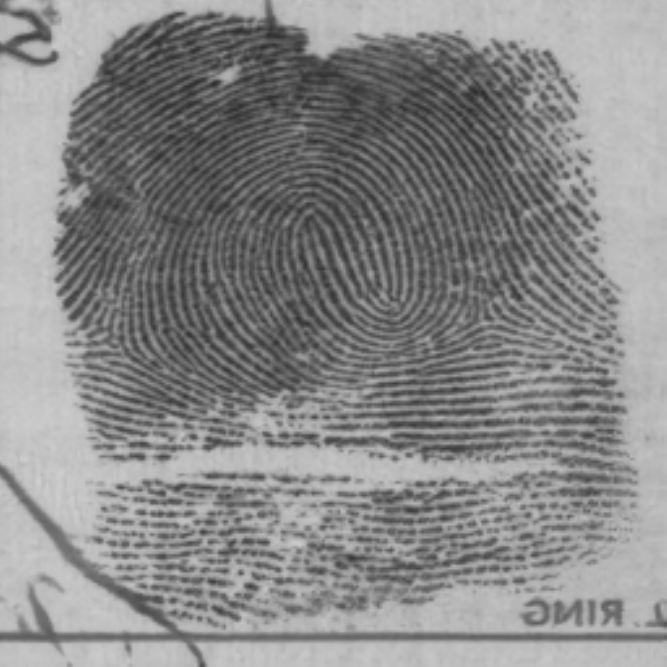} & %
\includegraphics[width=0.28\columnwidth,height=0.28%
\columnwidth]{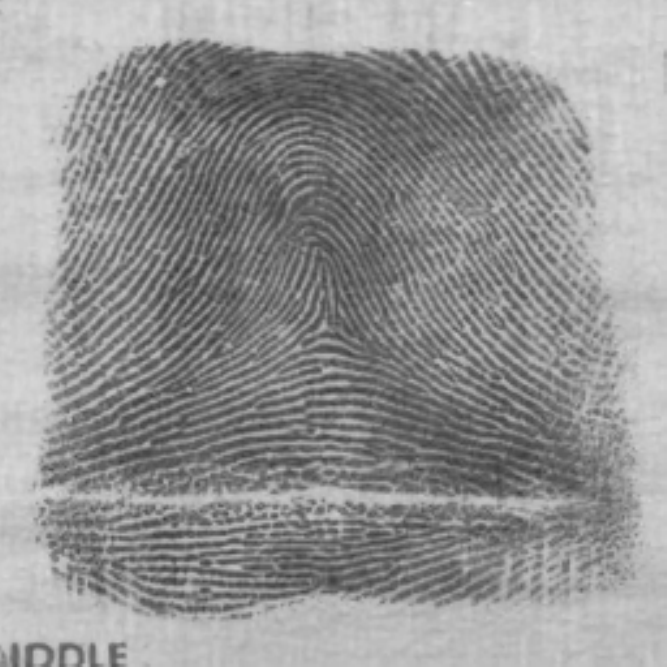} & %
\includegraphics[width=0.28\columnwidth,height=0.28%
\columnwidth]{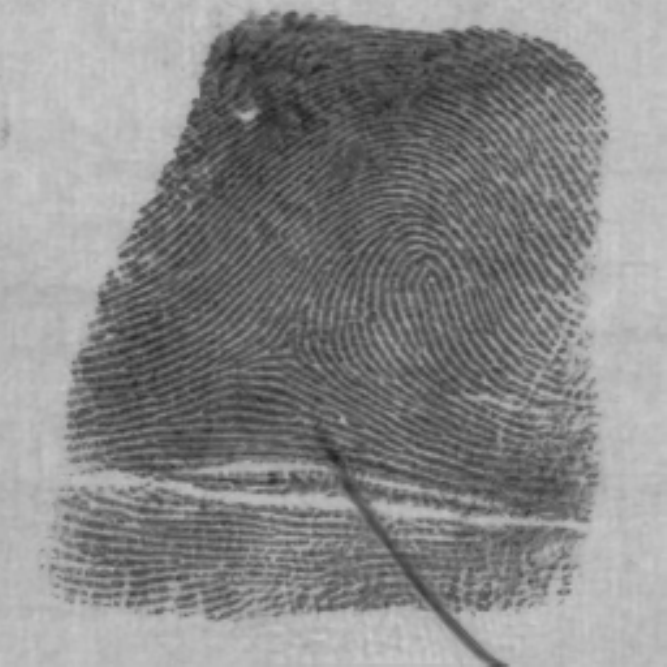} & %
\includegraphics[width=0.28\columnwidth,height=0.28%
\columnwidth]{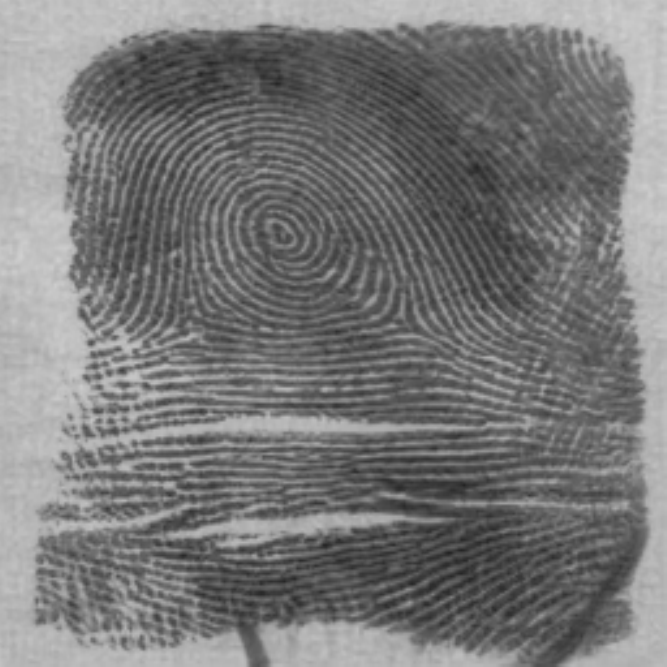} & %
\includegraphics[width=0.28\columnwidth,height=0.28%
\columnwidth]{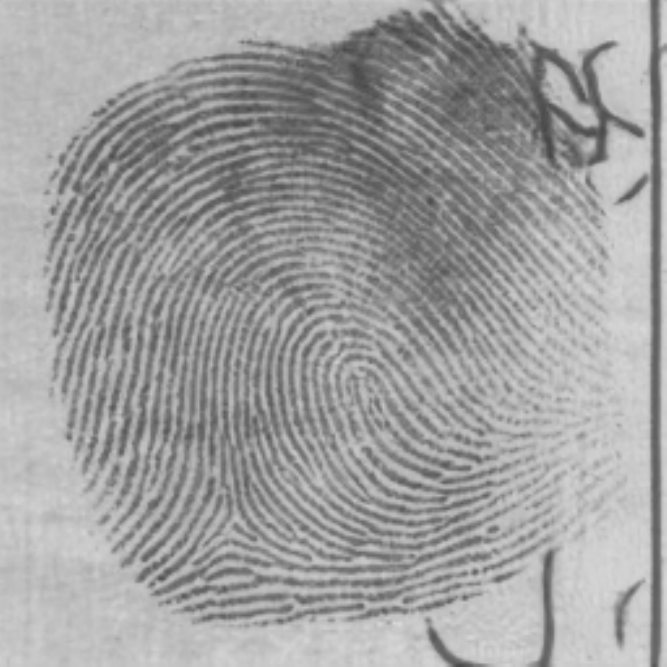} & %
\includegraphics[width=0.28\columnwidth,height=0.28%
\columnwidth]{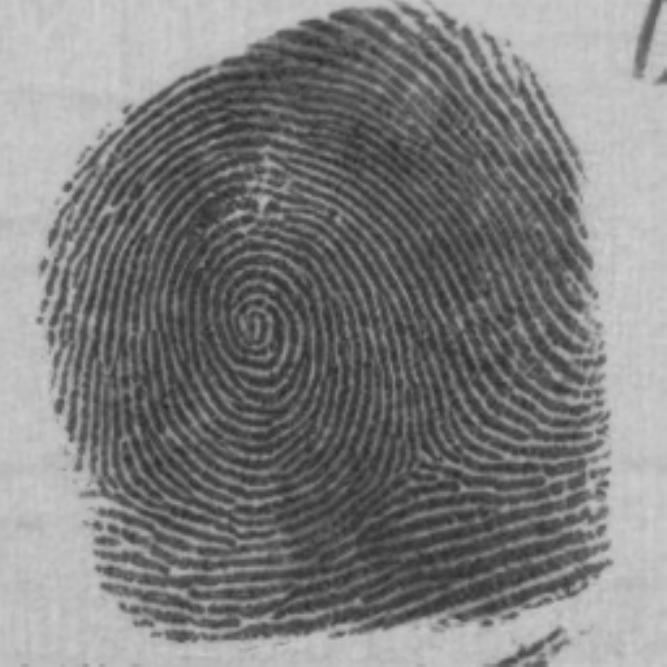} \\
\raisebox{+2.5\height}{\rotatebox[origin=b]{90}{CaoJain}} & %
\includegraphics[width=0.28\columnwidth,height=0.28%
\columnwidth]{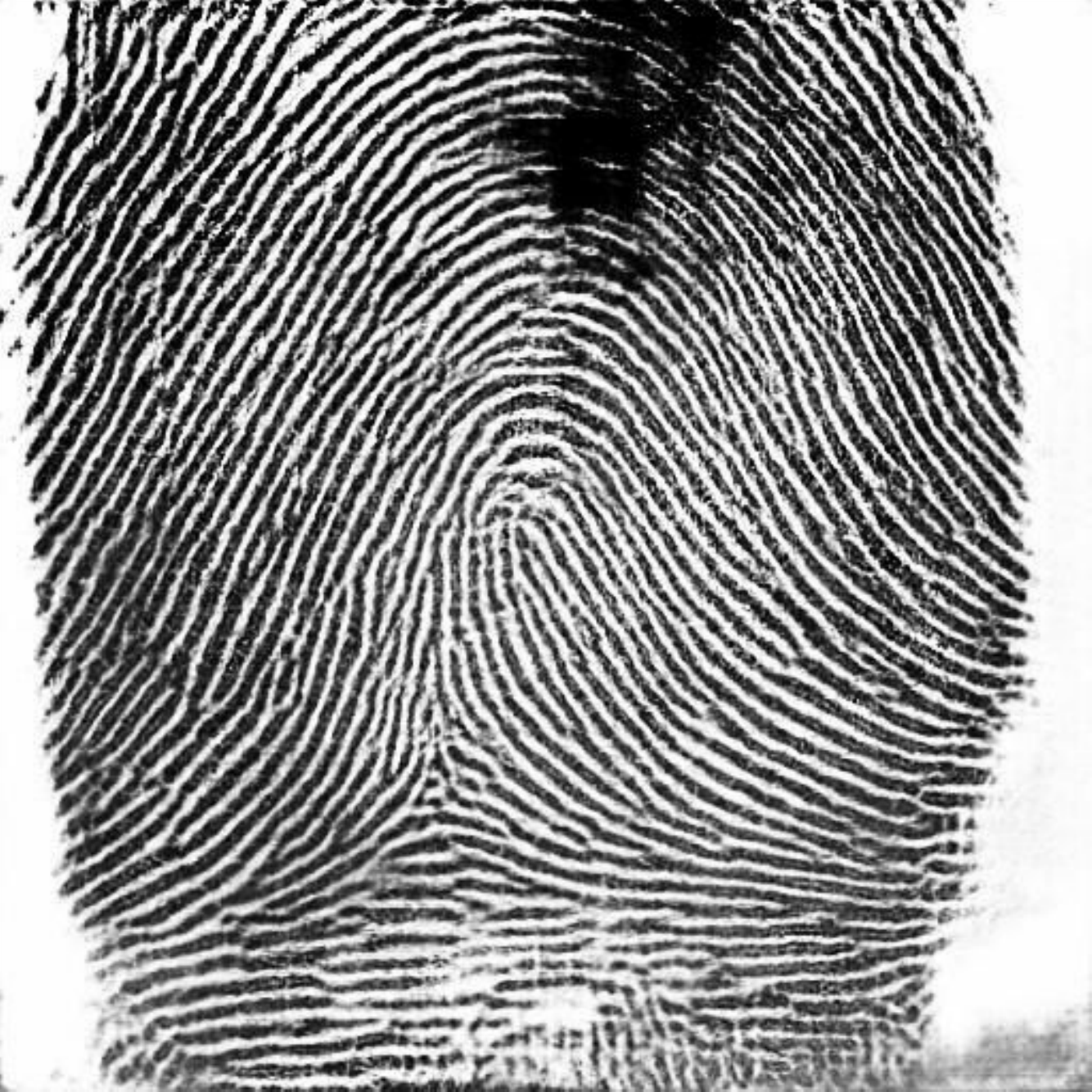} & %
\includegraphics[width=0.28\columnwidth,height=0.28%
\columnwidth]{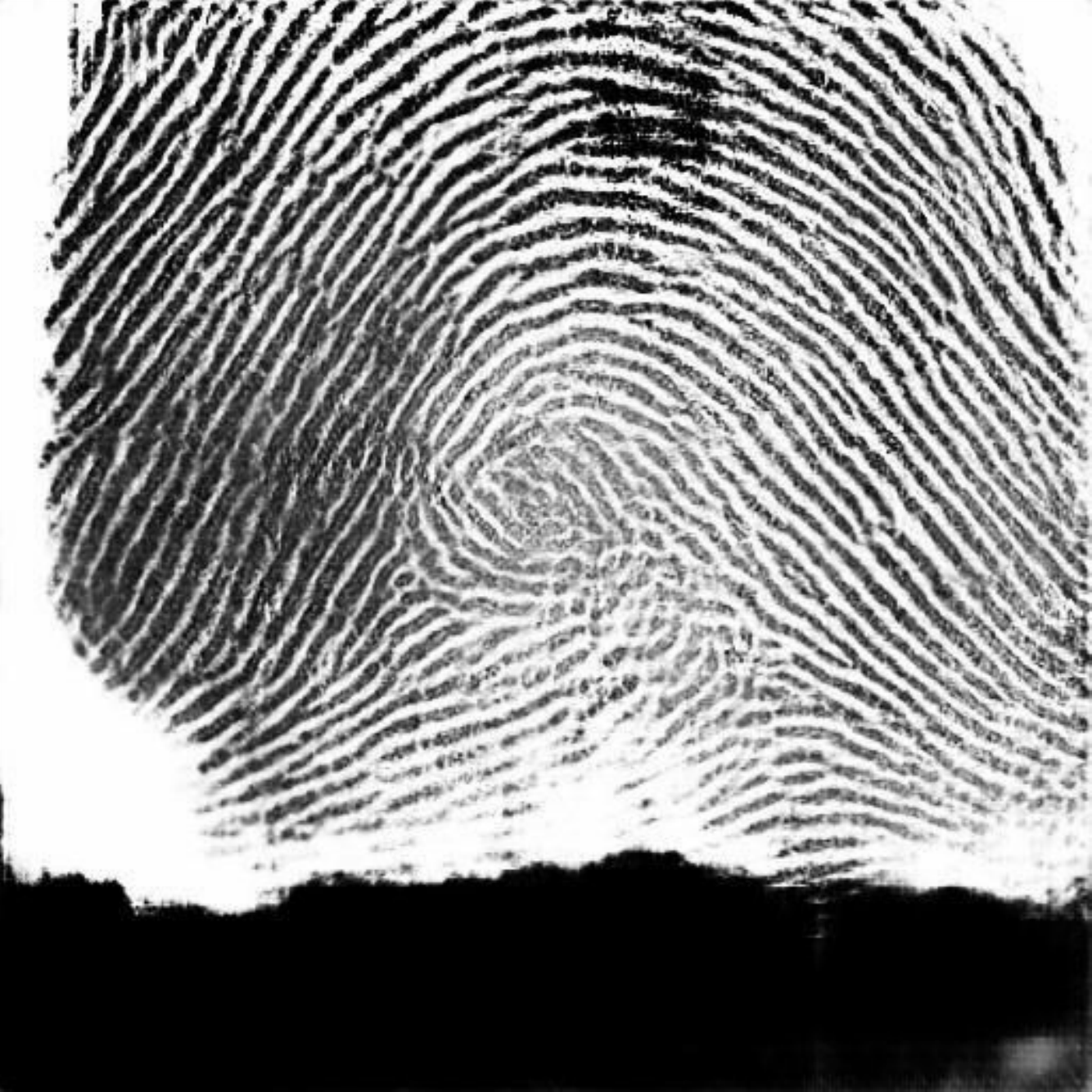} & %
\includegraphics[width=0.28\columnwidth,height=0.28%
\columnwidth]{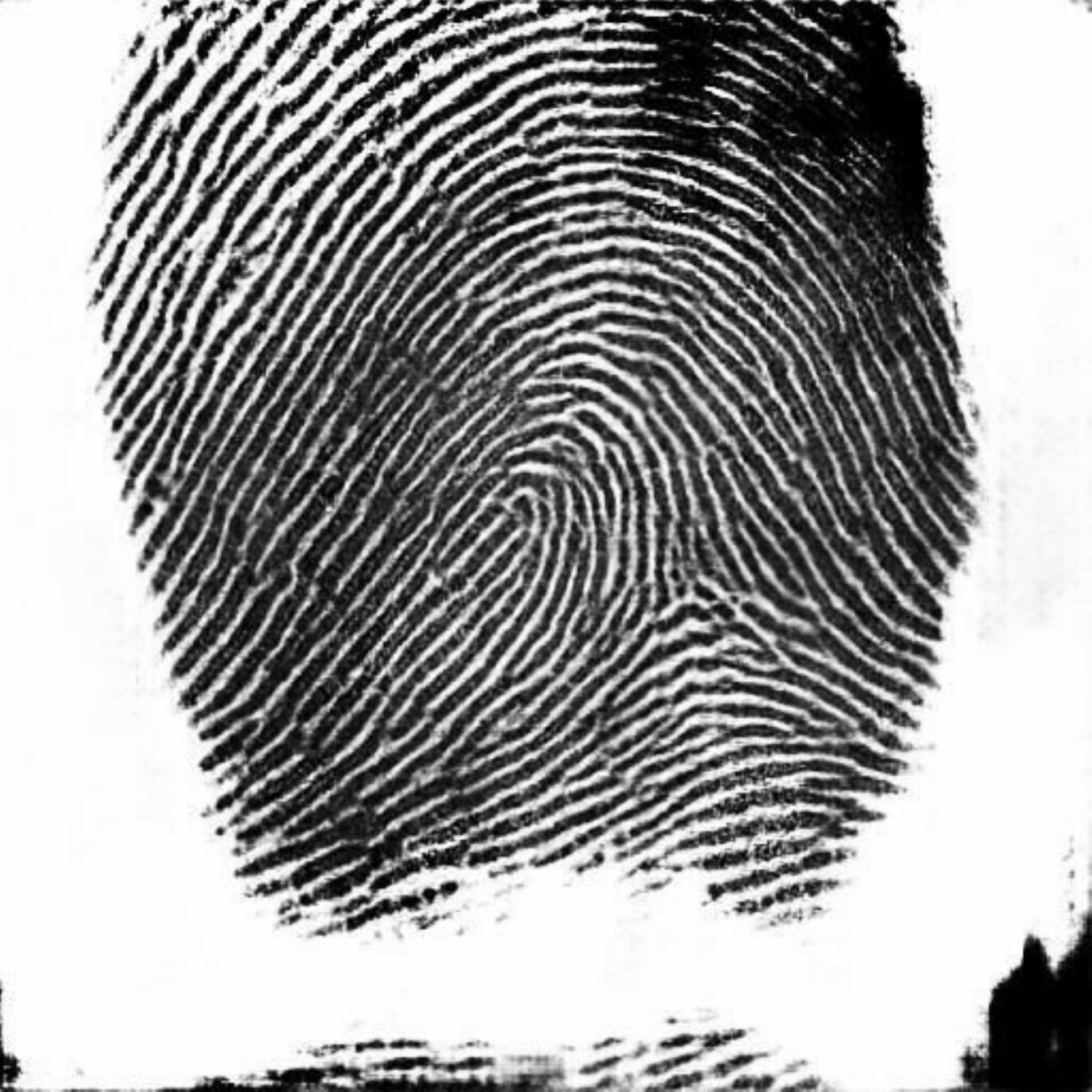} & %
\includegraphics[width=0.28\columnwidth,height=0.28%
\columnwidth]{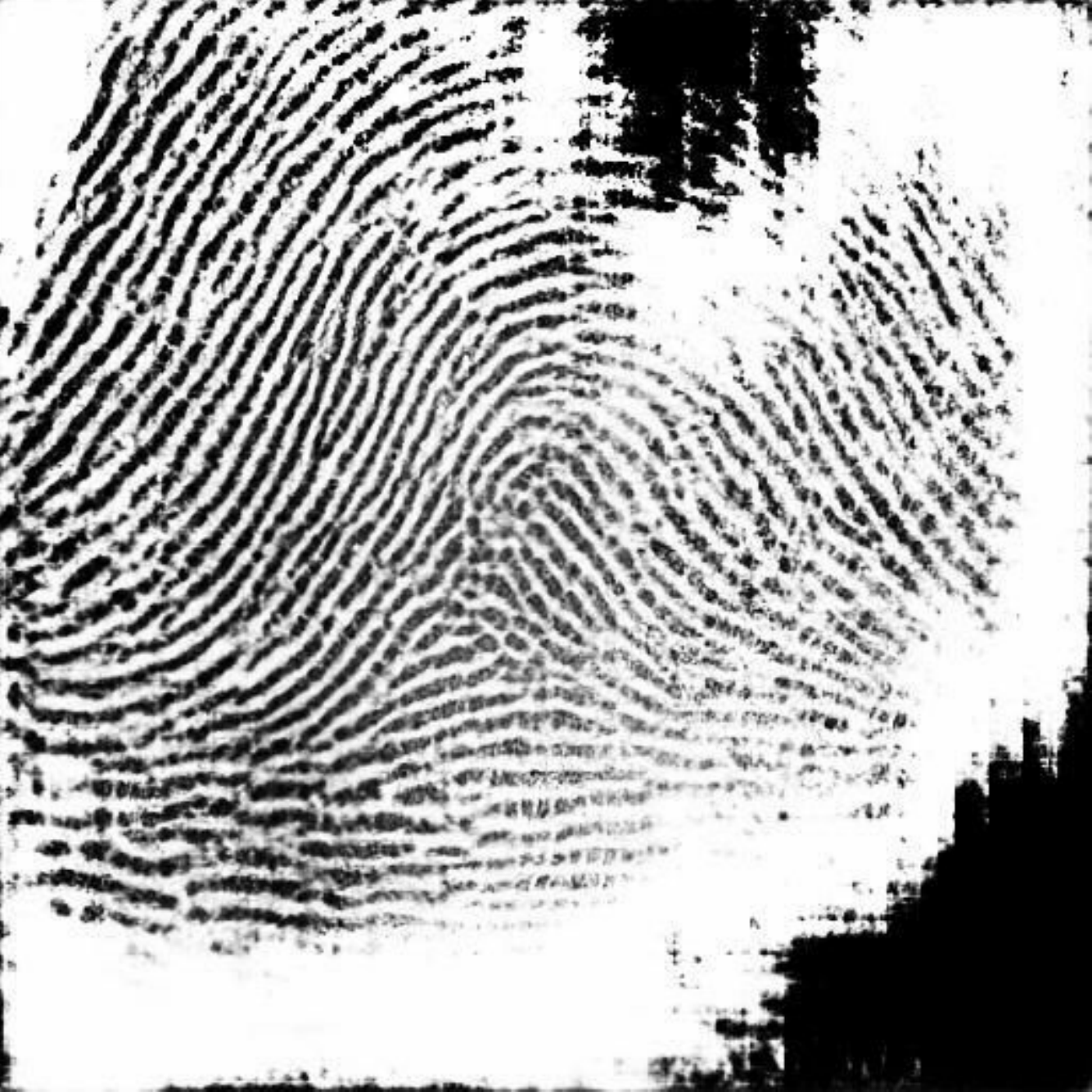} & %
\includegraphics[width=0.28\columnwidth,height=0.28%
\columnwidth]{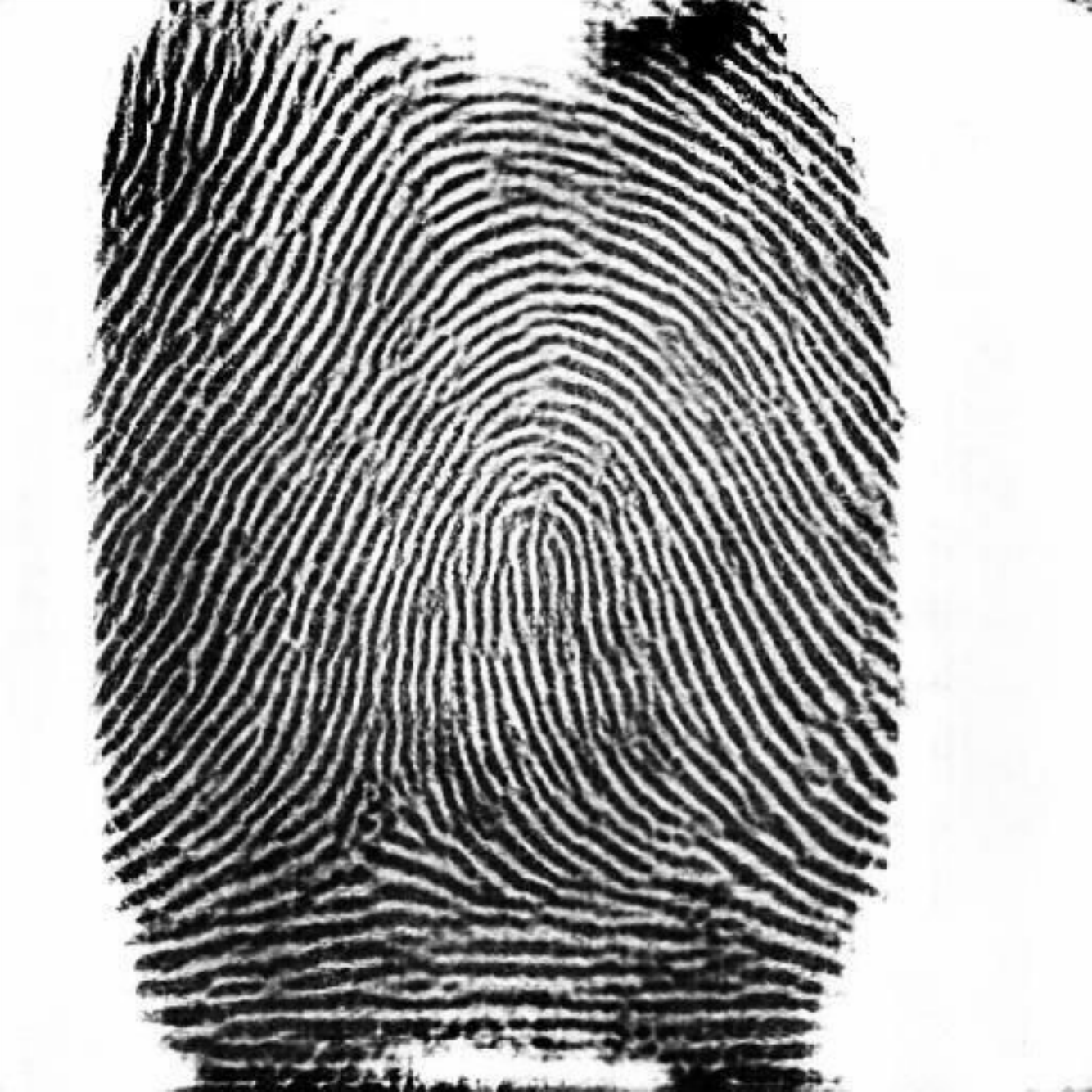} & %
\includegraphics[width=0.28\columnwidth,height=0.28%
\columnwidth]{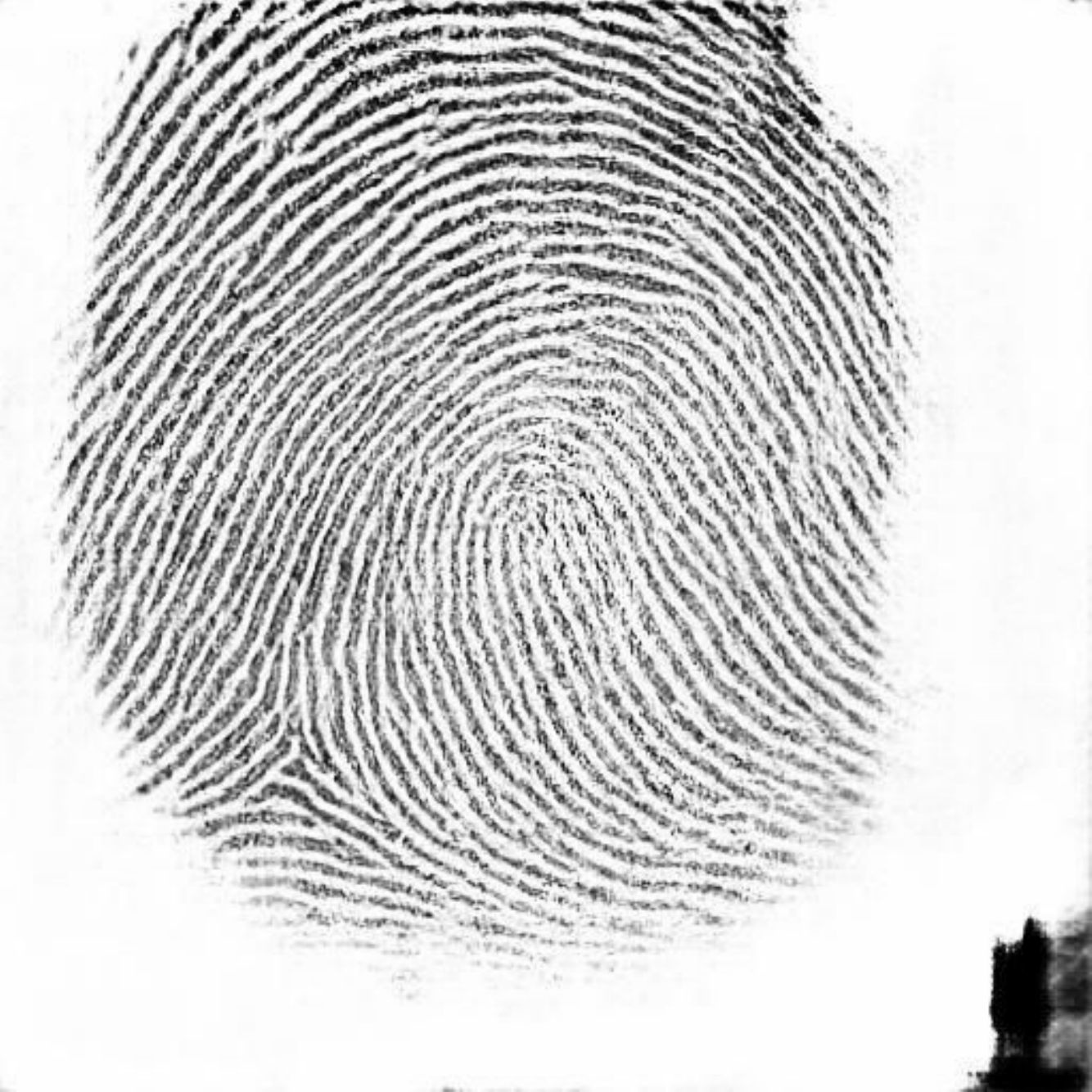}%
\end{tabular}%
\caption{\textbf{Visual Comparison} between NIST SD4 \protect\cite%
{watson1992nist}, SynFing and CaoJain \protect\cite{cao2018fingerprint}
datasets. The NIST SD4 and SynFing fingerprints are similar, in contrast to
the CaoJain samples that look artificial, and their attributes differ
significantly from those of NIST SD4.}
\label{fig:visual_examples}
\end{figure*}

To estimate the quality of the fingerprint generator, we evaluated the
fingerprint realism, distinctiveness, and distribution of minutiae
configurations., using the NIST SD4, SynFing, and CaoJain datasets. As NIST
SD4 is a dataset of real fingerprints, it estimates the \textquotedblleft
ideal\textquotedblright\ fingerprint synthesis method, which is an upper
bound of any fingerprint synthesis approach. Samples of each dataset are
shown in Fig.~\ref{fig:visual_examples}. Qualitatively, notable similarities
can be seen between the attributes of the NIST SD4 and our SynFing samples.
In contrast, the CaoJain images look artificial, and their visual attributes
differ significantly from those of the original ones.

\subsubsection{Fingerprints Validity}

For quantitative comparisons, we used the Fr\'{e}chet Inception Distance
(FID) \cite{heusel2017gans}, and the NIST Finger Image Quality (NFIQ 2.0)
\cite{nfiq2} scores. The FID is used to assess the quality of images created
by the generative model by comparing the distributions of synthesized and
real images. NFIQ 2.0 is an updated open source version of the widely used
NFIQ \cite{tabassi2004nist}, which computes a quality score given a
fingerprint image to predict the expected matching performance. The range of
NFIQ 2.0 scores is [0, 100], with 0 and 100 being the lowest and highest
quality scores, respectively.

Table~\ref{tab:FID} presents the FID score for both SynFing and CaoJain, as
well as the reported FID score for Minaee at el. \cite{minaee2018finger}.
The proposed scheme outperforms the previous schemes by a notable margin.
The mean and standard deviation of NFIQ 2.0 values for each of the mentioned
datasets are shown in Table~\ref{tab:NFIQ2_score}. The average NFIQ 2.0
values for NIST SD4, SynFing, and CaoJain are 44.7, 41.7 and 61.42,
respectively. The distribution of our proposed algorithm matches those of
the original dataset, while Cao and Jain's approach \cite{cao2018fingerprint}
exhibits significantly different distributions.
\begin{table}[th]
\centering{\
\begin{tabular}{ccc}
\toprule \textbf{Method} & \textbf{Dataset} & \textbf{FID score} \\
\multicolumn{1}{l}{Cao and Jain \cite{cao2018fingerprint}} & NIST SD4 &
113.82 \\
\multicolumn{1}{l}{Minaee at el. \cite{minaee2018finger}} & FVC 2006 & 70.55
\\
\multicolumn{1}{l}{Proposed framework\ \ } & NIST SD4 & \textbf{6.14} \\
\bottomrule &  &
\end{tabular}
}
\vspace{-0em}%
\caption{\textbf{Fr\'{e}chet inception distance (FID)} for different
fingerprint synthesis schemes (lower is better). We calculated the FIDs
using 40,000 random images generated by each scheme. The score of Minaee at
el. \protect\cite{minaee2018finger} is cited as there is no open-source
implementation available.}
\label{tab:FID}
\end{table}
\begin{table}[th]
\centering{\
\begin{tabular}{ccc}
\toprule \textbf{Dataset} & \textbf{Mean} & \textbf{Std. Dev.} \\
\multicolumn{1}{l}{NIST SD4} & 44.66 & 17.60 \\
\multicolumn{1}{l}{CaoJain} & 61.42 & 14.55 \\
\multicolumn{1}{l}{SynFing} & 41.70 & 16.10 \\
\bottomrule &  &
\end{tabular}
}
\vspace{-0em}%
\caption{\textbf{Distributions of NFIQ 2.0} NFIQ 2.0 values are [0,100],
where 0 and 100 indicate the lowest and highest quality values,
respectively. }
\label{tab:NFIQ2_score}
\end{table}

% \begin{figure}[th]
% \centering
% \includegraphics[width=\columnwidth]{fingerprint_image_quality.pdf} \vspace{%
% -1em}
% \caption{\textbf{Distributions of NFIQ 2.0} NFIQ 2.0 values are [0,100],
% where 0 and 100 indicate the lowest and highest quality values,
% respectively. }
% \label{fig:NFIQ2_score}
% \end{figure}

\subsubsection{Distinctiveness}

\begin{figure}[th]
\centering
\includegraphics[width=\columnwidth]{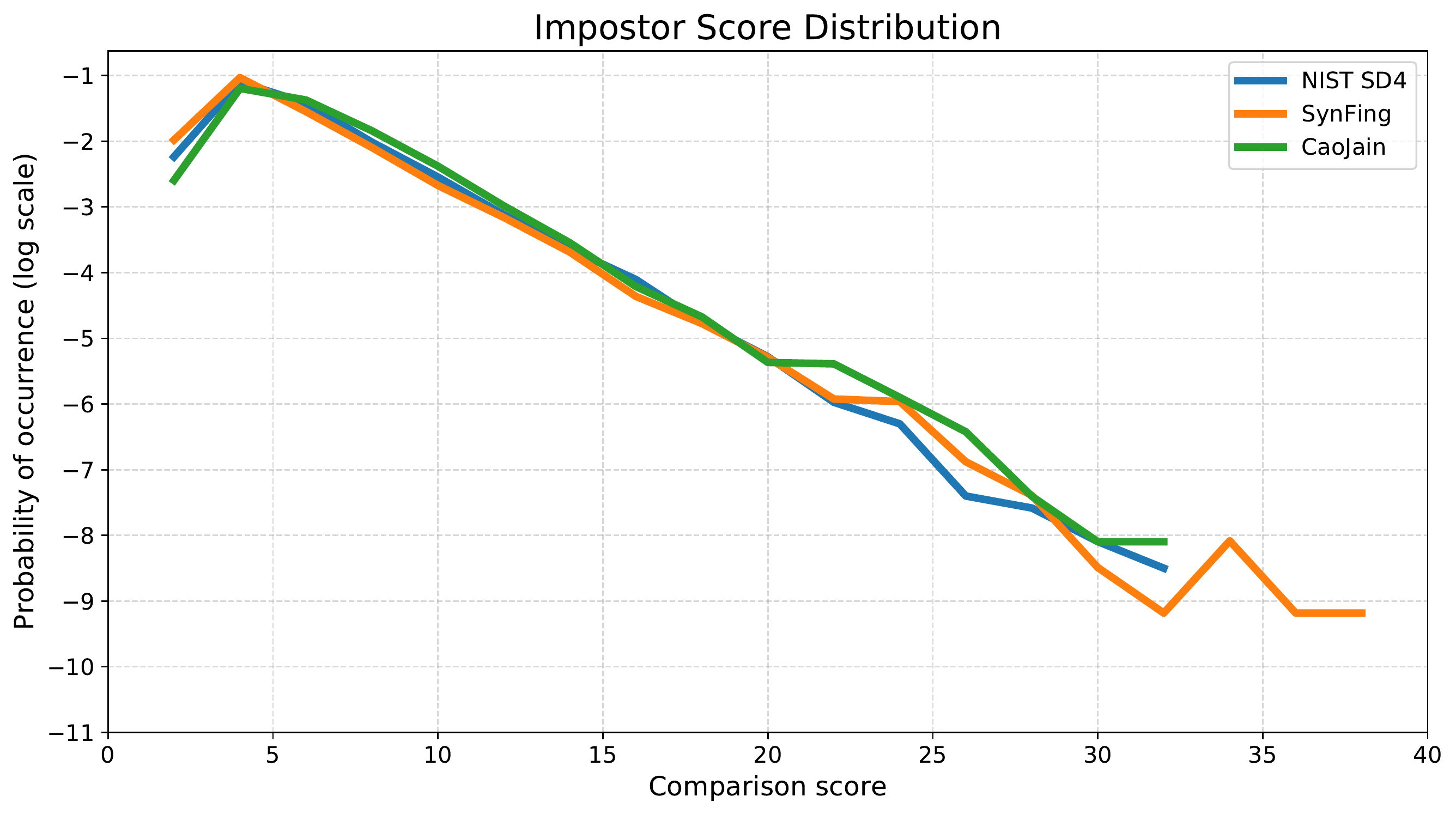}
\caption{\textbf{Impostor score distributions} for the NIST SD4, SynFing and
CaoJain datasets. The comparison scores of the Verifinger SDK 11.1 are
[0,6,356]. The higher the score, the more similar are the fingerprint
images. The probability of occurrence is shown in a log scale to illustrate
the differences for higher values.}
\label{fig:Impostor_score}
\end{figure}

To evaluate the diversity of our synthetic fingerprints (in terms of
identity), we compute pairwise comparison scores for each database using
Verifinger SDK 11.1. As long as the impostor score distribution is lower,
the generated fingerprints are more distinct. Figure~\ref{fig:Impostor_score}
shows the impostor score distributions. While these score distributions are
similar, they vary at the higher range of score values. The maximum impostor
comparison scores on NIST SD4, SynFing and CaoJain are 31, 31, and 36,
respectively. This indicates the higher diversity of the fingerprints
generated by our scheme (in terms of distinctiveness) compared to those
generated by the other approaches. Furthermore, this comparison reveals that
the number of corrupted fingerprints in CaoJain (identified by Verifinger as
'BadObject') is double that of those detected in SynFing (1.3\%).

\subsubsection{Minutiae Configuration}

\begin{figure}[th]
\centering
\subfigure[NIST
SD4]{\includegraphics[width=0.28\columnwidth,height=0.28\columnwidth]{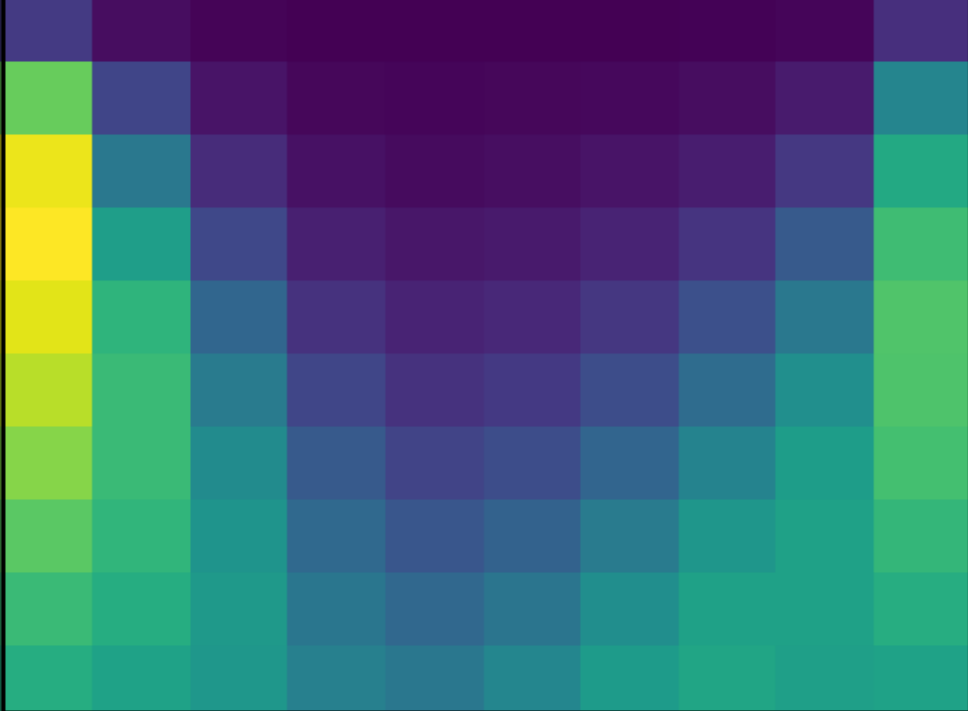}}
\subfigure[SynFing]{\includegraphics[width=0.28\columnwidth,height=0.28%
\columnwidth]{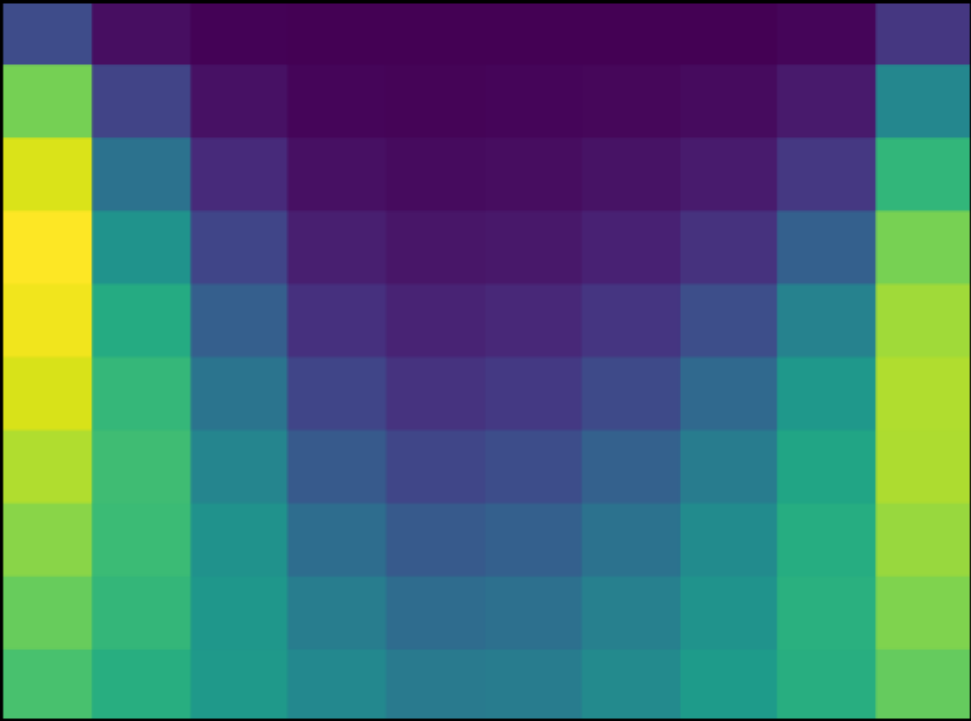}} \subfigure[CaoJain]{%
\includegraphics[width=0.28\columnwidth,height=0.28%
\columnwidth]{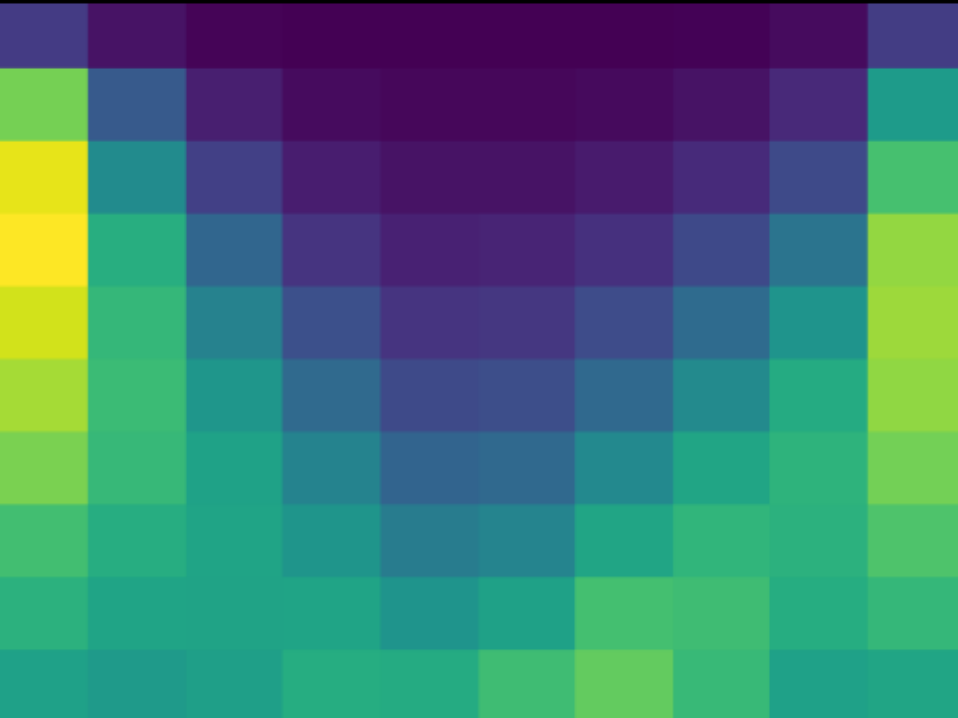}}
\caption{\textbf{Average 2D minutiae histograms}. The vertical axis of each
histogram represents the distance $d$ between minutiae locations, and the
horizontal axis shows the angular difference $\protect\alpha $ between
minutiae directions.}
\label{fig:2dmh}
\end{figure}

Fingerprint minutiae are considered the most discriminating features for
fingerprint recognition \cite{maltoni2009handbook}. The spatial distribution
of the configurations of the minutiae extracted from the synthesized
fingerprints is an indicator of their realism \cite%
{gottschlich2014separating}. Gottschlich and Huckemann \cite%
{gottschlich2014separating} showed that the 2D minutiae histogram (2DMH) is
effective in differentiating real fingerprint images from synthetic ones. We
compared the 2DMH of the SynFing and CaoJain datasets to that of SD4 to
evaluate the realism of synthetic fingerprint images. Given a fingerprint
image, its set of minutiae is extracted by Verifinger SDK 11.1. Then we
build a two-dimensional minutiae histogram by computing the distance $d$
between the minutiae locations (in pixels) and the angular difference $%
\alpha $ (in degrees) of the two minutiae directions for all pairs of
minutiae on a template. Both features are binned using identically sized
equidistant intervals. Figure~\ref{fig:2dmh} presents the average 2DMH for
SD4, SynFing, and CaoJain with $10\times 10$ bins, where the distances are
divided into intervals of 20 pixels, up to a maximum distance of 200 pixels
(distance increases from top to bottom). Angular differences are also
divided into 10 bins of $180^{\circ }$ overall. Each bin of angular
differences consists of two intervals of $18^{\circ }$ and the differences $>180^{\circ}$ are mirrored.

Figure~\ref{fig:2dmh} shows that the histograms of all the three datasets
are qualitatively similar. Therefore, we computed the Earth Moving Distance
(EMD) between the histograms, following Gottschlich and Huckemann \cite%
{gottschlich2014separating}, to quantitatively compare the different
methods. The EMD measured between NIST SD4 and SynFing is 0.42, while that
between NIST SD4 and CaoJain is 0.73, indicating that the fingerprint images
generated by our approach (SynFing) are more similar to real fingerprint
images (NIST SD4) than those of the previous SOTA approach (CaoJain).

\subsection{Fingerprints Reconstruction}

\label{sec:Fingerprint Reconstruction}

Reconstructed fingerprints can be used to spoof a system that stores the
original fingerprint templates (referred to as a Type-I attack), or other
systems where the same finger has been enrolled with a different impression
(referred to as a Type-II attack). We evaluated the proposed reconstruction
scheme for verification and identification, using the VeriFinger 11.1
fingerprint recognition system and the NIST SD4 dataset.

\subsubsection{Fingerprint Verification}

The verification experiment was carried out using NIST SD4 and Type-I and
Type-II attacks. In a Type-I attack, each reconstructed fingerprint is
matched against the same impression from which the minutiae template was
extracted, while in a Type-II attack, each reconstructed fingerprint is
matched against another impression of the same finger, which is more
difficult. Based on imposter matching scores, thresholds for different false
acceptance rates (FAR) are calculated. In Table \ref{tab:verification}, we
report the verification performances of the Type-I and Type-II attacks. As
none of the fingerprint reconstruction algorithms are publicly available and
as there are no verification results for NIST SD4, we only report the
results of the proposed scheme.
\begin{table}[th]
\centering{{\
\begin{tabular}{ccc}
\bottomrule \textbf{FAR} & \textbf{Type-I attack} & \textbf{Type-II attack}
\\
\multicolumn{1}{l}{1\%} & 99.92\% & 97.67\% \\
\multicolumn{1}{l}{0.1\%} & 99.67\% & 93.67\% \\
\multicolumn{1}{l}{0.01\%} & 99.26\% & 86.32\% \\
\multicolumn{1}{l}{0\%} & 99.23\% & 85.44\% \\
\bottomrule &  &
\end{tabular}%
} }%
\vspace{-0em}%
\caption{\textbf{Verification accuracy} of the proposed approach for type-I
and type-II attacks and multiple false accept rates (FAR) using the NIST SD4
dataset.}
\label{tab:verification}
\end{table}

In the Type-I attack, the matching performance of all four tested FAR are
above 99\%, implying that in over 99\% of cases the proposed framework was
able to reconstruct the original fingerprint at a level that\textit{\ would
allow the deception of a commercial fingerprint recognition system}. For the
Type-II attack, our results illustrate the ability of the proposed algorithm
to reconstruct a fingerprint image while preserving its original attributes.

\subsubsection{Fingerprint Identification}

In the fingerprint identification experiments, minutiae templates of 2,000
file fingerprints from NIST SD4 were used to reconstruct the fingerprints.
Each reconstructed fingerprint is compared to 2,000 file fingerprints to
obtain 2,000 Type-I attacks, and to 2,000 query fingerprints to obtain 2,000
Type-II attacks. Table~\ref{tab:identification} reports the identification
performance for Type-I and Type-II attacks of the proposed approach, as well
as the performance of the four other reconstruction schemes by Cao and Jain
\cite{cao2014learning}, Feng and Jain \cite{feng2010fingerprint}, Li and Kot
\cite{li2012improved}, and Ross \cite{ross2007template}.

\begin{table}[th]
\centering{\
\begin{tabular}{ccc}
\toprule \textbf{Method} & \textbf{Type-I attack} & \textbf{Type-II attack}
\\
\multicolumn{1}{l}{Ross \cite{ross2007template}} & 23.00\% & - \\
\multicolumn{1}{l}{Li and Kot \cite{li2012improved}} & 90.80\% & 24.80\% \\
\multicolumn{1}{l}{Feng and Jain \cite{feng2010fingerprint}} & 99.70\% &
65.10\% \\
\multicolumn{1}{l}{Cao and Jain \cite{cao2014learning}} & 99.05\% & 71.00\%
\\
\multicolumn{1}{l}{Proposed framework\ \ } & \textbf{99.89\%} & \textbf{%
98.93\%} \\
\bottomrule &  &
\end{tabular}
}
\vspace{-0em}%
\caption{\textbf{The identification accuracy} of fingerprints reconstruction
schemes for type-I and type-II attacks using the NIST SD4 dataset. The
results of Li and Kot \protect\cite{li2012improved} are cited \protect\cite%
{cao2014learning} as their identification results were not reported.}
\label{tab:identification}
\end{table}

The identification accuracy of the proposed reconstruction algorithm is
significantly better than that of other reconstruction schemes. For the
Type-I attack, the proposed scheme shows almost perfect results. In fact, in
99.9\% of cases, the source fingerprint is identified as the best match
among all the gallery of 2000 fingerprints. In terms of Type II attack,
which is considered more challenging, the proposed algorithm outperforms all
other approaches by showing an identification performance of 98.9\% compared
to 71.0\% of Cao and Jain \cite{cao2014learning}, which was the previous
SOTA in this task.

\subsection{Fingerprint Attribute Modification}

Our approach allows the user to generate multiple impressions of the same
fingerprint by modifying the attributes of the generated fingerprint, as
detailed in Section \ref{subsec:AttributesModification}. To evaluate the
performance, we used the proposed SynFing synthetic fingerprint dataset that
was split into two subsets. Each subset consists of fingerprints generated
by modifying one of the two leading interpretable directions found by
investigating the latent space of the proposed generator using the SeFa
approach \cite{shen2021closedform}. The first direction affects the presence
of scribbles in the fingerprint background, while the second adds blobs and
dry-skin artifacts to the generated fingerprint. We refer to these datasets
as SynFingP1 and SynFingP2, respectively, which are shown in Fig.~\ref%
{fig:Sefa_examples} and were made publicly available.
\begin{figure}[th]
\centering\subfigure[Backward]{\includegraphics[width=0.33%
\columnwidth,height=0.33\columnwidth]{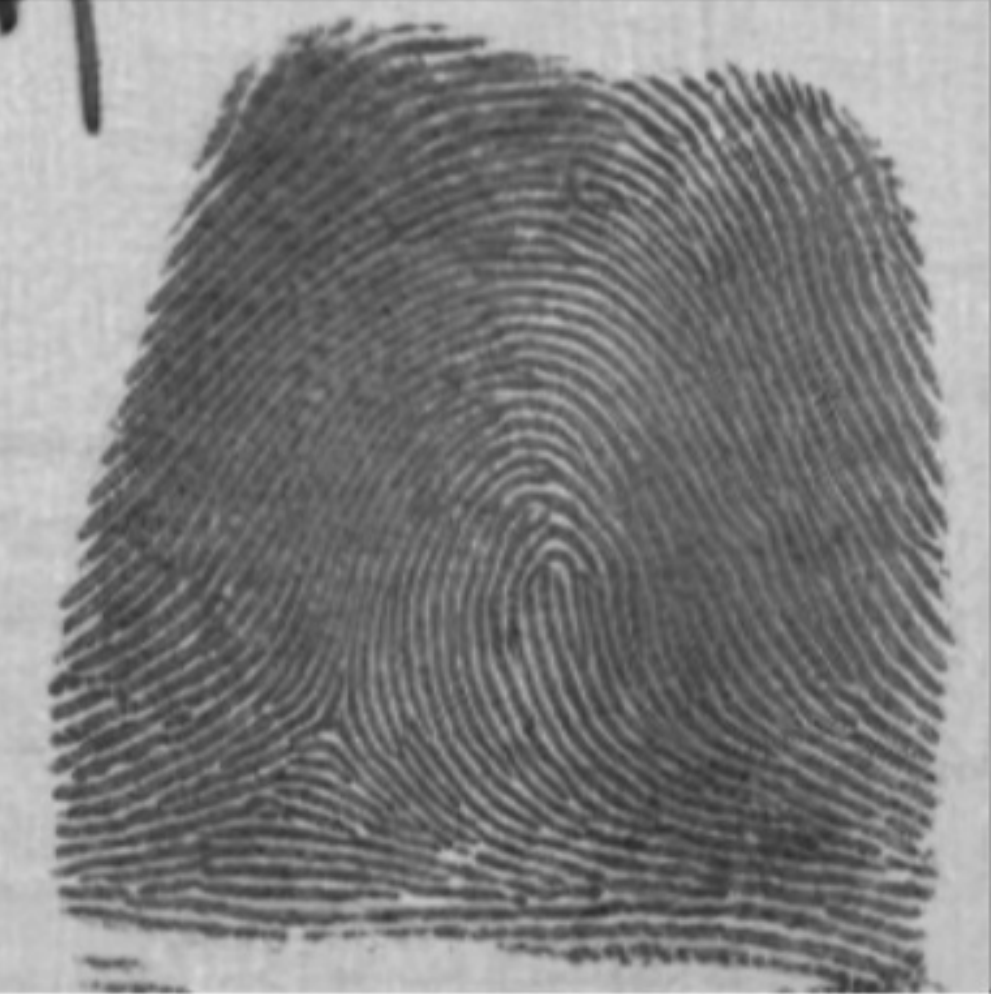}}%
\subfigure[Original]{\includegraphics[width=0.33\columnwidth,height=0.33%
\columnwidth]{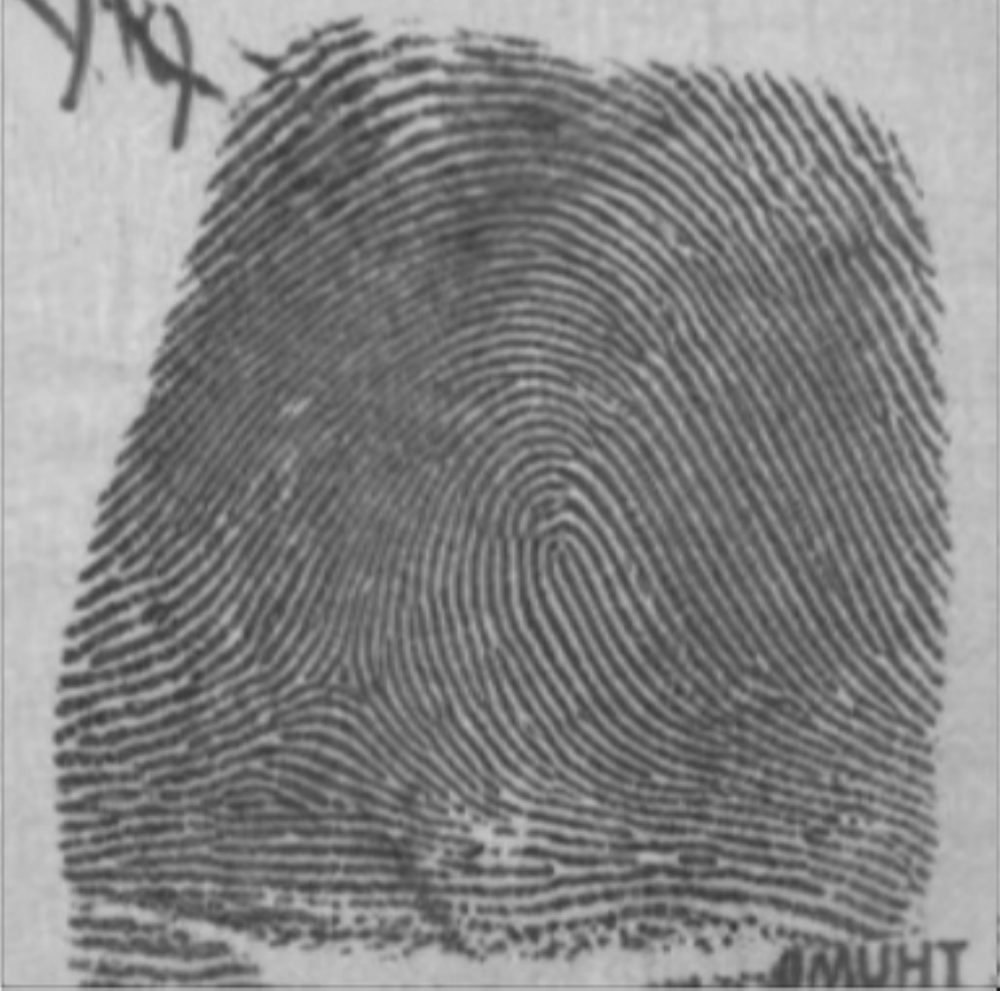}}\subfigure[Forward]{%
\includegraphics[width=0.33\columnwidth,height=0.33%
\columnwidth]{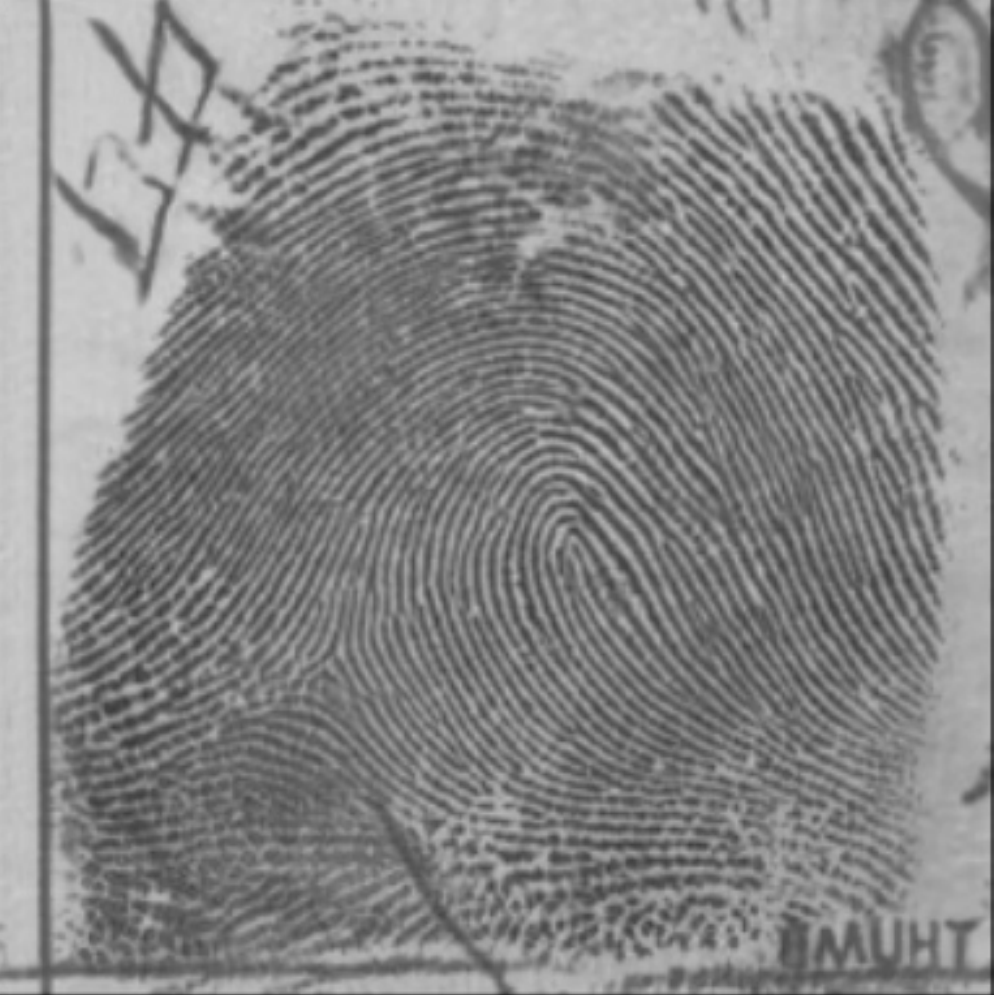}} \newline
\subfigure[Backward]{\includegraphics[width=0.33\columnwidth,height=0.33%
\columnwidth]{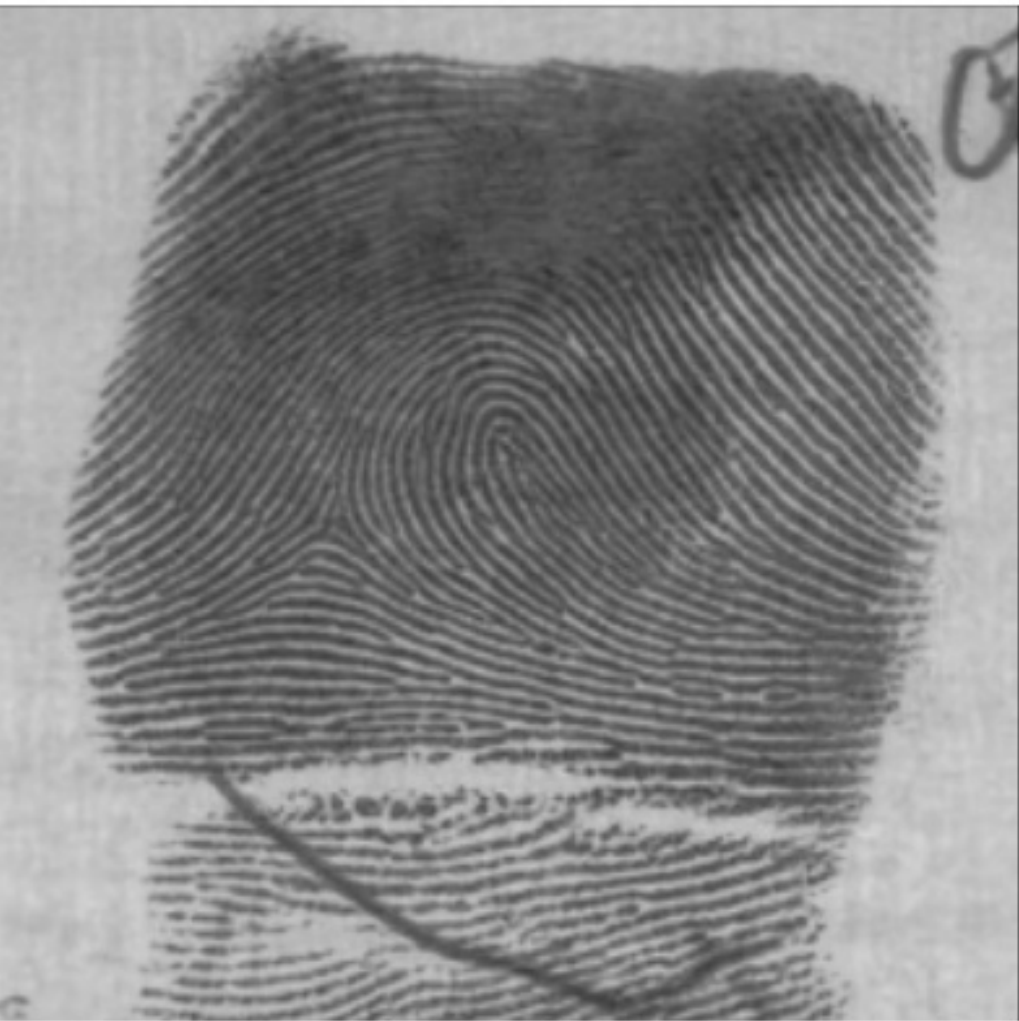}}\subfigure[Original]{%
\includegraphics[width=0.33\columnwidth,height=0.33%
\columnwidth]{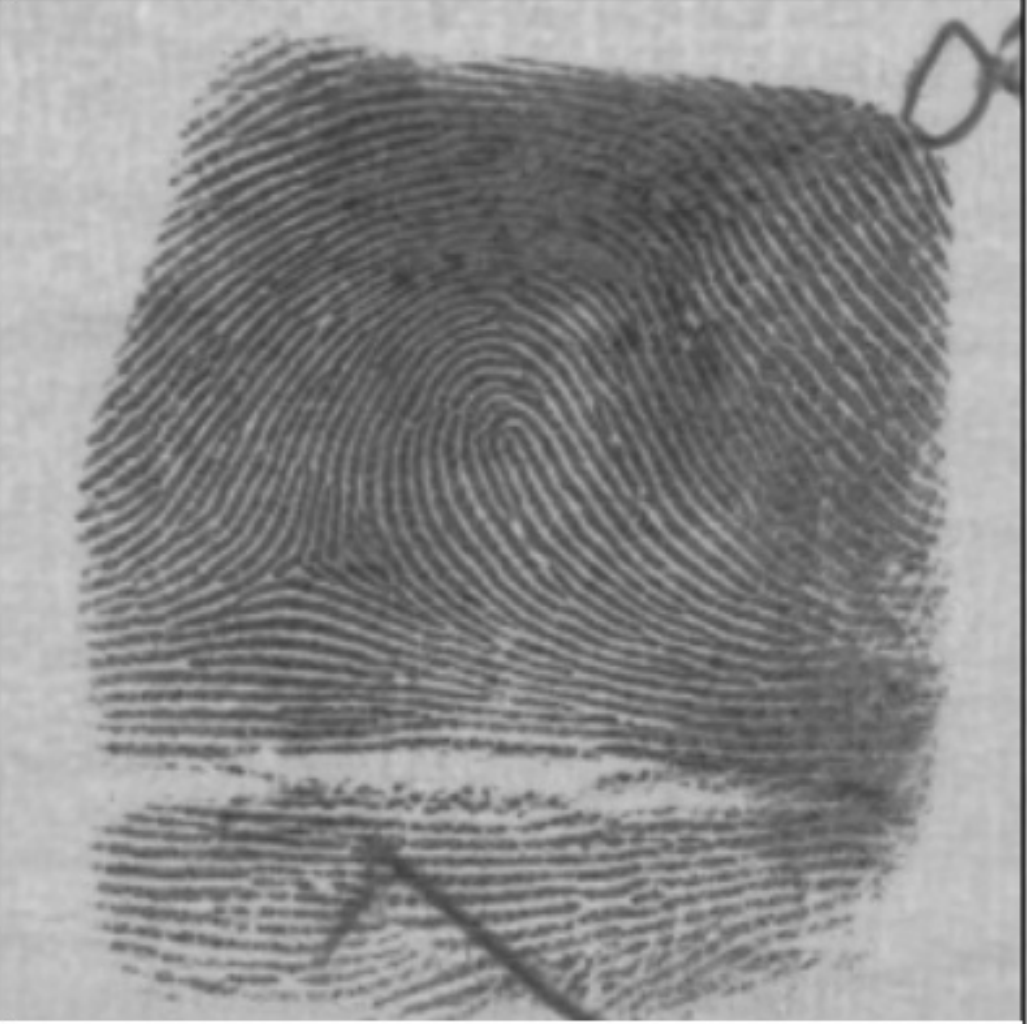}}\subfigure[Forward]{%
\includegraphics[width=0.33\columnwidth,height=0.33%
\columnwidth]{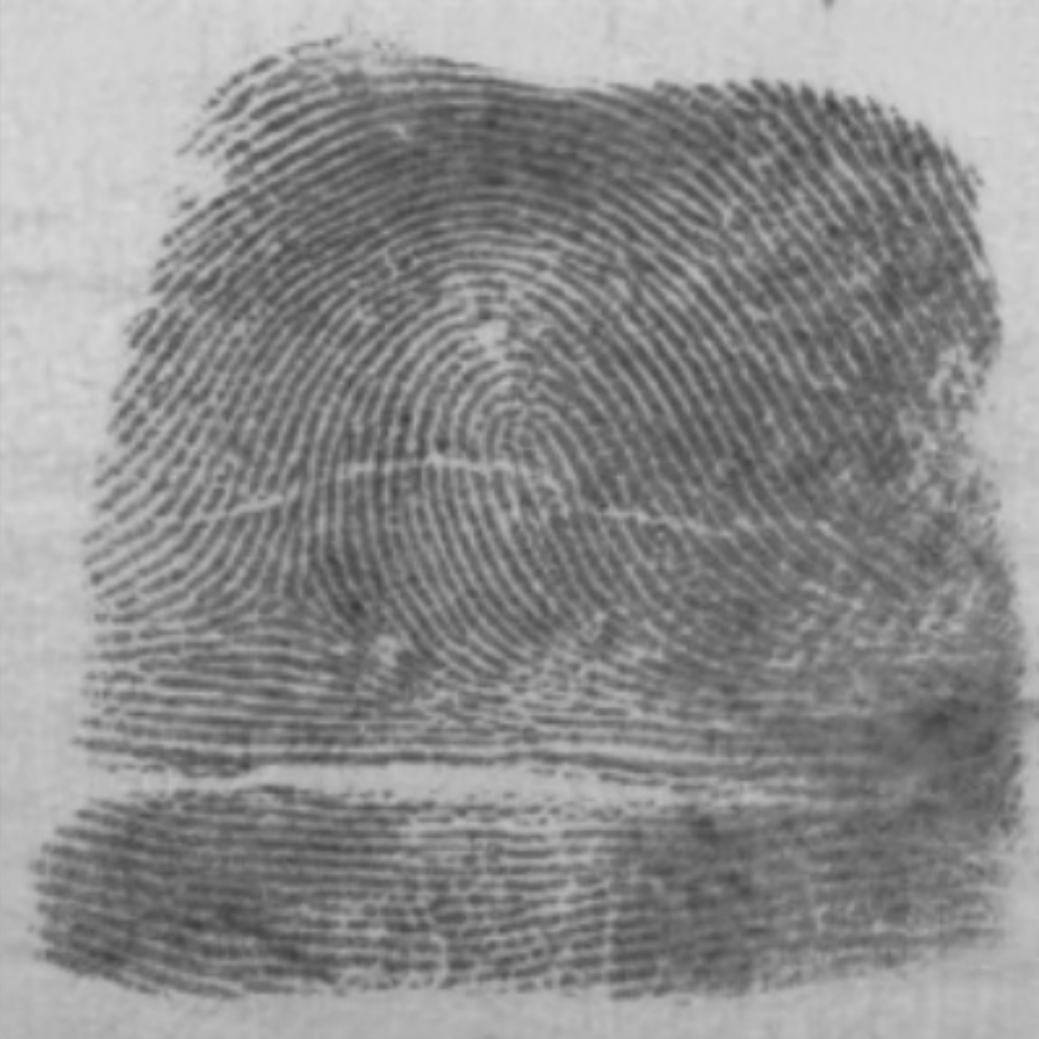}}
\caption{\textbf{The leading directions in the fingerprint embedding space.}
\textbf{Top row:} SynFingP1, the direction affects the scribbles in the
fingerprint background. \textbf{Bottom row:} SynFingP2, the direction,
modifies the dry-skin artifacts in the generated fingerprint. The middle
image is the original reconstruction, while the other images are the results
of modifying the embedding forward and backward in the leading direction.}
\label{fig:Sefa_examples}
\end{figure}

\begin{figure}[th]
\centering
\includegraphics[width=\columnwidth]{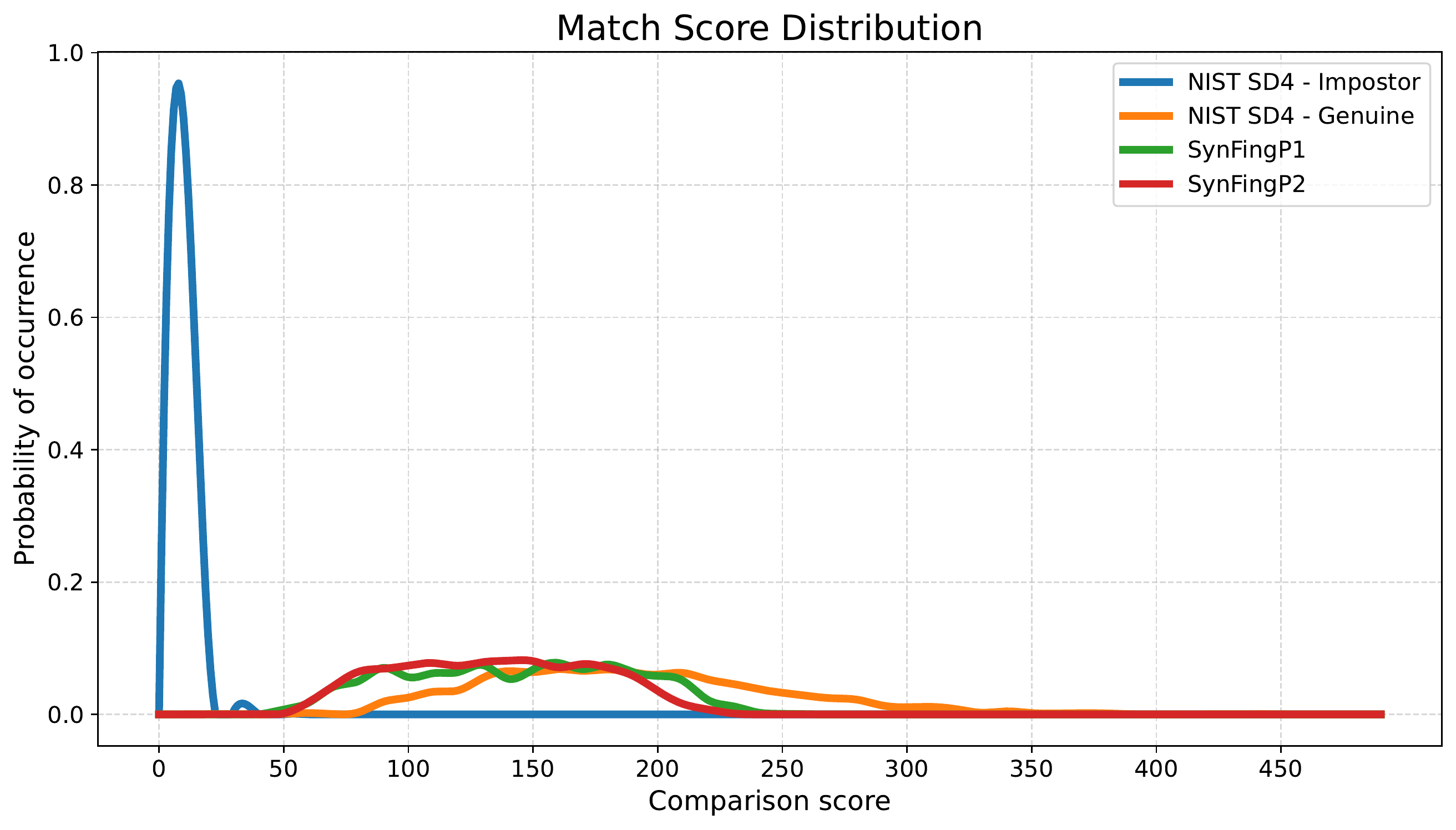}
\caption{\textbf{Matching score distributions} of the NIST SD4, SynFingP1
and SynFingP2 datasets. The higher the score, the more similar are the
fingerprint images. The score distribution for the SD4 dataset relates to
impostor comparison, while for SynFingP1 and SynFingP2 it relates to the
matching accuracy.}
\label{fig:Sefa_metching_scores}
\end{figure}

Figure~\ref{fig:Sefa_examples} shows the varying attributes across multiple
impressions, and we quantitatively evaluate the identity preservation across
all impressions. Therefore, for each pair of impressions, we compute the
matching scores using Verifinger SDK 11.1, and compare them to the pairwise
matching scores for random fingerprints from the NIST SD4 dataset. Figure~%
\ref{fig:Sefa_metching_scores} shows the matching score distributions for
each dataset. For the NIST SD4 dataset, the impostor and genuine distributions are the same,
while those of SynFingP1 and SynFingP2 are only a genuine comparison. The
matching scores for SynFingP1 and SynFingP2 are significantly higher than
NIST SD4 impostor score and slightly less compared to the genuine score.
This implies that the identity of a fingerprint is well-preserved across the
generated impressions, yet there is still room for improvement in order to
achieve results similar to those of the real fingerprints.

\subsection{Improving Deep Network Training}

One of the main motivations for which we have designed our fingerprint
generator is the ability to train a deep learning model using the synthetic
dataset to solve a variety of tasks in the field of fingerprints. To
evaluate the effectiveness of using our synthetic dataset, we trained three
fingerprint matching models, based on the ArcFace architecture \cite%
{deng2019arcface}. Each model is trained using a different training set,
NIST SD14, SynFing, and the combination of the two. The last one was
pretrained with the SynFing dataset and then fine-tuned with the NIST SD14
dataset.

The verification accuracy of the three models in the NIST SD4 dataset is
detailed in Table~\ref{tab:fingerprint_matchers}. The model trained with the
SynFing dataset is 2\% more accurate than the model trained using only the
NIST SD14 dataset, while the model trained with both datasets outperforms
the two other models by a significant margin. These results demonstrate the
effectiveness of using the proposed synthetic fingerprint dataset to augment
the limited real fingerprint datasets.

\begin{table}[th]
\centering{\
\begin{tabular}{ccc}
\toprule \textbf{Training Set} & \textbf{Verification Accuracy} &  \\
\multicolumn{1}{l}{NIST SD14} & 69.20\% &  \\
\multicolumn{1}{l}{SynFing} & 71.43\% &  \\
\multicolumn{1}{l}{SynFing + NIST SD14} & 83.72\% &  \\
\bottomrule &  &
\end{tabular}
}
\vspace{-0em}%
\caption{\textbf{Verification Accuracy} of three verification models trained on a
real fingerprint dataset, a synthetic fingerprint dataset, and the
combination of them.}
\label{tab:fingerprint_matchers}
\end{table}

\section{Conclusions}

\label{sec:Conclusions}

We introduced a new framework in this study for fingerprint synthesis and
reconstruction that employs generative adversarial networks. Our approach
employs the StyleGan2 architecture as the fingerprint generator for both
tasks. We presented a Minutiae-To-Vec encoder that encodes minutia into
latent vectors for fingerprint reconstruction and a novel fingerprint
synthesis approach that manipulates the attributes of the generated
fingerprints while preserving their identity. Our proposed scheme is
experimentally shown to outperform state-of-the-art methods for both
fingerprint synthesis and reconstruction. This has enhanced the realism of
the generated fingerprints, both visually and in terms of spoofing
fingerprint-based systems. Additionally, we have shared the SynFing dataset
of 100,000 synthetic fingerprint pairs.

\bibliographystyle{IEEEtran}
\bibliography{bibliographyList}

\end{document}